\makeatletter
\def\input@path{{arXiv-2605.19597v1/}}
\makeatother
\PassOptionsToPackage{framemethod=tikz}{mdframed}
\documentclass[11pt,letterpaper,onecolumn,logo]{hunyuan_arxiv}
\graphicspath{{}{arXiv-2605.19597v1/}}

\usepackage[authoryear,sort&compress,round]{natbib}
\usepackage[most,breakable,skins]{tcolorbox}
\tcbuselibrary{skins,breakable}
\usepackage{microtype}
\usepackage{wrapfig}

\let\cite\citep

\usepackage{amsmath,amsfonts,amssymb,amsthm}
\usepackage{booktabs,array,tabularx,multirow,ragged2e,xcolor}
\usepackage{tikz}
\usetikzlibrary{arrows.meta,positioning,shapes.geometric,fit,backgrounds,calc}
\usepackage[most]{tcolorbox}

\definecolor{exposeRed}{HTML}{B91C1C}
\definecolor{exploitAmber}{HTML}{B7791F}
\definecolor{misleadBlue}{HTML}{1D4ED8}
\definecolor{judgeBlue}{HTML}{1D4ED8}
\definecolor{softGray}{HTML}{F3F4F6}
\definecolor{borderGray}{HTML}{CBD5E1}
\definecolor{codeBg}{HTML}{F8FAFC}
\definecolor{codeFrame}{HTML}{D5DDE8}
\definecolor{codeTitleBg}{HTML}{EEF4FF}
\definecolor{codeTitleText}{HTML}{1E3A8A}
\definecolor{codeAccent}{HTML}{2563EB}
\definecolor{tableHeadBg}{HTML}{EAF1F8}
\definecolor{tableBandBg}{HTML}{F7FAFC}
\definecolor{tableGroupBg}{HTML}{F1F5F9}
\definecolor{tableRule}{HTML}{94A3B8}
\definecolor{tableText}{HTML}{1F2937}

\usepackage{enumitem}
\setlist[itemize]{leftmargin=*}
\setlist[enumerate]{leftmargin=*}

\theoremstyle{definition}

\AtBeginEnvironment{tabular}{\arrayrulecolor{tableRule}}
\AtBeginEnvironment{tabularx}{\arrayrulecolor{tableRule}}

\newtcblisting{boxedcode}[1][]{
  enhanced,
  listing only,
  listing engine=listings,
  breakable,
  width=0.88\textwidth,
  center,
  colback=codeBg,
  colframe=codeFrame,
  colbacktitle=codeTitleBg,
  coltitle=codeTitleText,
  fonttitle=\sffamily\bfseries\footnotesize,
  boxrule=0.45pt,
  arc=2pt,
  borderline west={2pt}{0pt}{codeAccent},
  left=9pt,
  right=9pt,
  top=7pt,
  bottom=7pt,
  before skip=8pt,
  after skip=10pt,
  listing options={
    basicstyle=\ttfamily\scriptsize\color{black!82},
    breaklines=true,
    columns=fullflexible,
    keepspaces=true,
    showstringspaces=false
  },
  #1
}

\AtBeginDocument{%
  \setlength{\parindent}{0pt}%
  \setlength{\parskip}{0.6\baselineskip plus 2pt}%
}

\makeatletter
\renewcommand{\abscontent}{}
\renewenvironment{abstract}
  {\begin{tcolorbox}[
      colback=gblue9!5,
      colframe=gblue9!5,
      boxrule=0pt,
      arc=6pt,
      left=8pt,right=8pt,
      top=6pt,bottom=6pt,
      breakable]
   \absfont}
  {\end{tcolorbox}\par\bigskip}
\makeatother

\usepackage{amsmath,amsfonts,bm}

\def\eqref#1{equation~\ref{#1}}

\def\1{\bm{1}}

\DeclareMathAlphabet{\mathsfit}{\encodingdefault}{\sfdefault}{m}{sl}
\SetMathAlphabet{\mathsfit}{bold}{\encodingdefault}{\sfdefault}{bx}{n}

\usepackage{fvextra}
\usepackage{hyperref}
\usepackage{longtable}
\usepackage{fontawesome5}
\usepackage{microtype}
\usepackage{booktabs}
\usepackage{pifont} 
\usepackage{multirow}
\usepackage{xspace}
\usepackage{color}
\usepackage{adjustbox}
\usepackage[edges]{forest}
\usepackage{amssymb}  
\usepackage{amsfonts}
\usepackage{tabularx}
\usepackage{tikz}
\usepackage{makecell}
\usepackage{caption}
\usepackage{subcaption}
\usepackage{graphicx}
\usepackage{amsmath}
\usepackage[utf8]{inputenc}
\usepackage[T1]{fontenc}
\usepackage{CJKutf8}

\DeclareUnicodeCharacter{2018}{`}%
\DeclareUnicodeCharacter{2019}{'}%
\DeclareUnicodeCharacter{201C}{``}%
\DeclareUnicodeCharacter{201D}{''}%
\DeclareUnicodeCharacter{2013}{--}%
\DeclareUnicodeCharacter{2014}{---}%
\DeclareUnicodeCharacter{2011}{-}%
\DeclareUnicodeCharacter{2026}{\ldots}%
\DeclareUnicodeCharacter{2022}{\textbullet}%
\DeclareUnicodeCharacter{25CF}{\textbullet}%
\DeclareUnicodeCharacter{2192}{\ensuremath{\rightarrow}}%
\DeclareUnicodeCharacter{00B7}{\textperiodcentered}%
\DeclareUnicodeCharacter{00BB}{\guillemotright}%
\DeclareUnicodeCharacter{00D7}{\ensuremath{\times}}%
\DeclareUnicodeCharacter{00B0}{\textdegree}%
\DeclareUnicodeCharacter{00B1}{\ensuremath{\pm}}%
\DeclareUnicodeCharacter{00F7}{\ensuremath{\div}}%
\DeclareUnicodeCharacter{00BA}{\textordmasculine}%
\DeclareUnicodeCharacter{00E9}{\'e}%
\DeclareUnicodeCharacter{200C}{}%

\usepackage{url}
\usepackage{nicefrac}
\usepackage{xcolor}
\usepackage[normalem]{ulem}
\usepackage[most]{tcolorbox}
\usepackage{colortbl}
\usepackage{tcolorbox}
\usepackage[tikz]{bclogo}
\usepackage{wrapfig}
\usepackage{float}
\usepackage{enumitem}
\usepackage{ifthen}
\ifdefined\directlua
  \IfFileExists{emoji.sty}{\usepackage{emoji}}{}%
\fi
\ifdefined\emoji\else
  \newcommand{\emojipenguin}{%
    \raisebox{-0.15\height}{\includegraphics[height=1.0em]{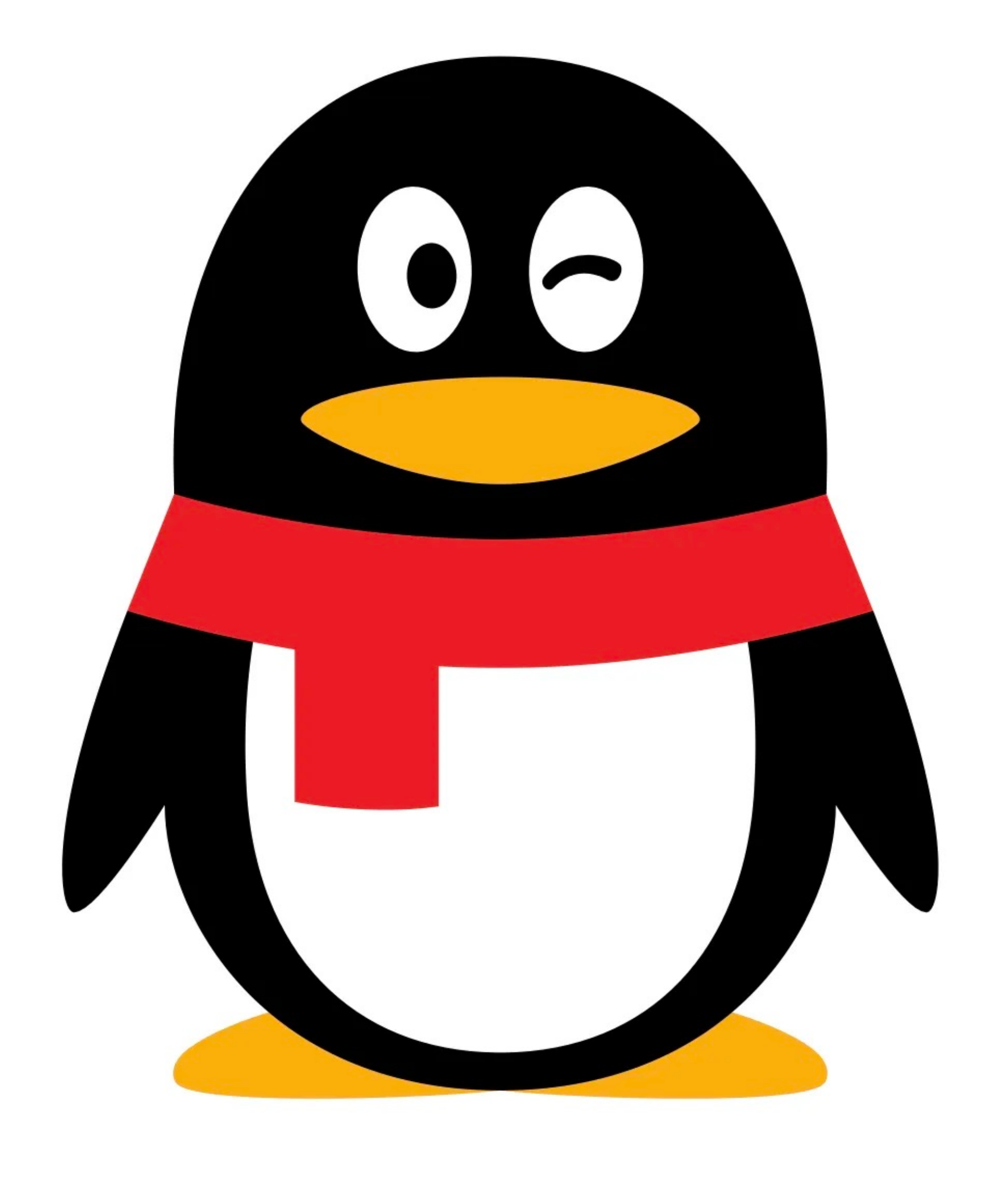}}%
  }%
  \providecommand{\emoji}[1]{\ifthenelse{\equal{#1}{penguin}}{\emojipenguin}{}}%
\fi

\usepackage{listings}
\usepackage{xparse}
\ExplSyntaxOn

\NewDocumentCommand{\gradientcell}{m}{
    \fp_set:Nn \l_tmpa_fp {#1}
    \fp_compare:nTF {\l_tmpa_fp > 20} {
        \cellcolor{green!60!yellow!\fp_eval:n{(\l_tmpa_fp-20)*2}!white}#1
    }{
        \fp_compare:nTF {\l_tmpa_fp > 10} {
            \cellcolor{yellow!60!red!\fp_eval:n{(\l_tmpa_fp-10)*3}!white}#1
        }{
            \cellcolor{red!\fp_eval:n{\l_tmpa_fp*5}!white}#1
        }
    }
}
\ExplSyntaxOff

\definecolor{deepred}{RGB}{230,101,101}

\definecolor{mygrey}{RGB}{105,105,105}

\usetikzlibrary{arrows.meta,positioning,fit,backgrounds,calc}

\renewcommand{\thefootnote}{\fnsymbol{footnote}}

\title{\emoji{penguin}-Bench: Benchmarking Multi-Step Tool-Use Agents in Real-World Product Scenarios}

\makeatletter
\def\AB@authnote#1{}
\def\AB@affilnote#1{}
\renewcommand\AB@affilsepx{\protect\\ \protect\Affilfont}
\makeatother
\author{
Weihuang Zheng$^{*1}$ \quad Tianyuan Zou$^{*2,1}$ \quad Eileen Ye$^1$ \quad Alphet Liu$^1$ \quad Youyong Kong$^3$ \\\mbox{Ya-Qin Zhang$^2$} \quad Duran Zheng$^1$ \quad Maxm Pan$^1$
}
\affil{$^1$Hunyuan Team, Tencent; \quad $^2$Institute for AI Industry Research, Tsinghua University;}
\affil{$^3$School of Computer Science and Engineering, Southeast University}

\begin{document}

\maketitle

\vspace{-1em}
\begin{abstract}

Large Language Models (LLMs) are increasingly deployed as agents that interact with stateful environments over multiple steps: gathering hidden information, composing tool calls, and committing state changes. We refer to this capability as \textbf{multi-step tool use}. Existing benchmarks have advanced tool-use agent evaluation, but often focus on isolated API calls, short trajectories, or settings that are difficult to scale or control. We introduce \verb|E-Bench|, a \emph{\textbf{fully synthetic}} benchmark with $323$ state-changing tasks across three product domains: \textit{Honor of Kings}, \textit{QQ Music}, and \textit{Tencent Meeting}. \verb|E-Bench| decouples \emph{environment synthesis} from \emph{task synthesis}: graph-guided database filling builds reusable, orphan-free product environments, while generator-solver asymmetry creates tasks with both an \emph{information gap} and a \emph{tool gap}, requiring agents to discover hidden data and compose multiple tool calls before changing state. Outcomes are graded deterministically by database-state diffs. Since both environments and tasks are synthetic, \verb|E-Bench| is controllable at the environment level and scalable at the task level. Benchmarking $11$ cutting-edge LLMs shows that multi-step tool use remains challenging: Pass$^3$ stays below $60\%$ for the strongest models, and even with code execution in \verb|E-Bench|-Code extension, reliability (Pass$^3$) remains below $70\%$.

\end{abstract}

\footnotetext[1]{
Equal contribution.
$^{\dagger}$Correspondence to maxmpan@tencent.com. 
To prevent benchmark-specific training or overfitting, we do not open-source the \texttt{E-Bench} environments or tasks.
}

\setcounter{footnote}{0}
\renewcommand{\thefootnote}{\arabic{footnote}}

\begin{figure}[H]
\centering
\vspace{-1.2em}
    \includegraphics[width=0.98\textwidth]{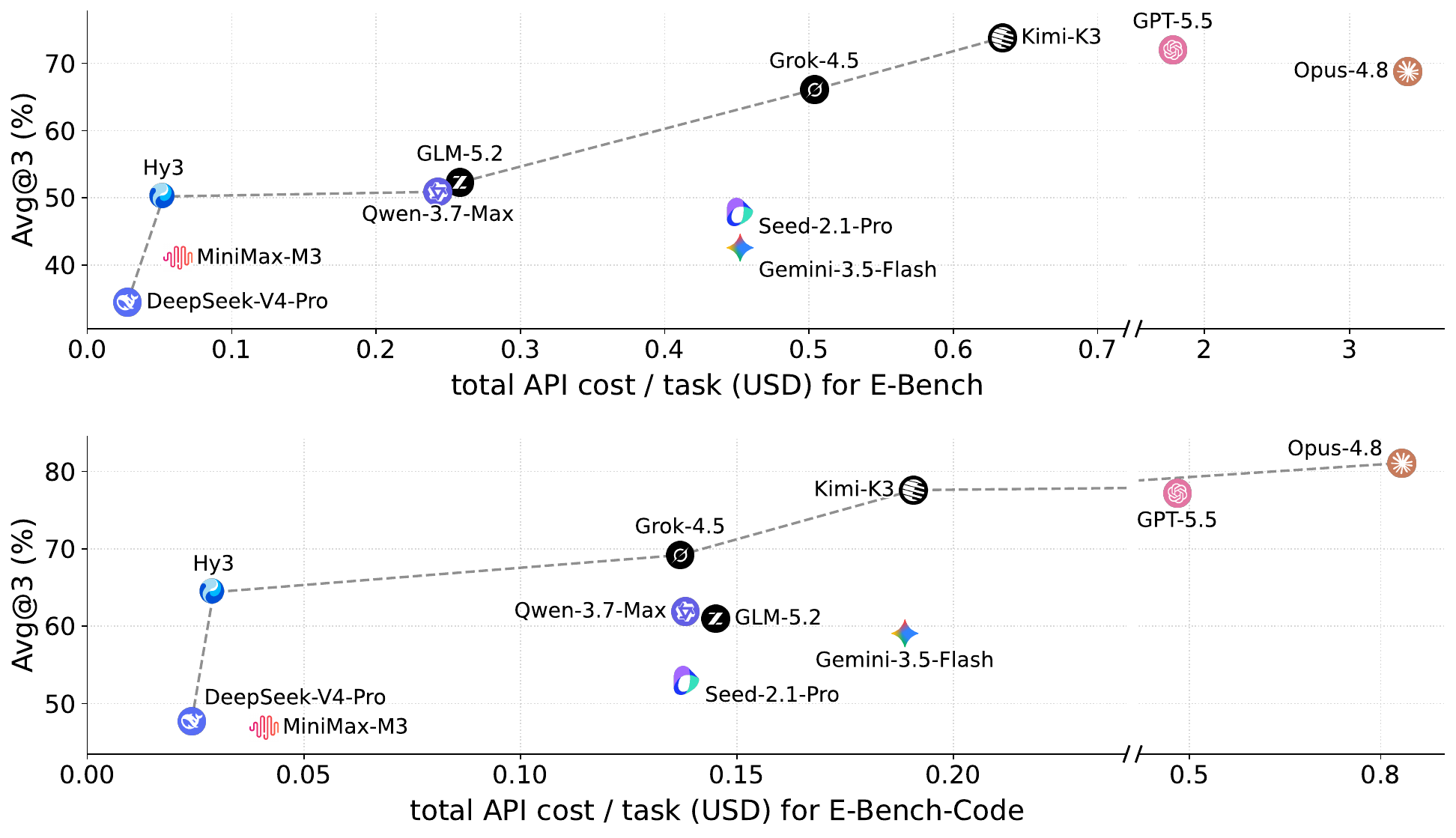}
\caption{
Avg@3 (\%) versus total API cost per task (USD)
under \texttt{E-Bench} (top) and \texttt{E-Bench}-Code (bottom). Calculation details are included in the Appendix~\ref{subsec:cost_appendix}.
The dashed line marks the Pareto (cost-efficiency) frontier. 
}
\label{fig:price_vs_performance}
\vspace{-0.5em}
\end{figure}

\section{Introduction}

Currently, Large Language Models (LLMs) have moved beyond answering self-contained prompts and are increasingly deployed as agents that interact with external environments to accomplish complex tasks~\cite{yao2022react, patil2024gorilla, singh2025openai, anthropic2024claude}. Success then requires more than a plausible response: an LLM agent must \emph{repeatedly} identify missing information, decide which tools to invoke, integrate observations across steps, and commit changes back to a stateful environment (we term this \textbf{multi-step tool use}). This shift from answering to acting calls for benchmarks that evaluate agentic behavior under realistic interaction constraints, rather than isolated language understanding or single-step tool selection.

\begin{figure}[!b]
\centering
\captionsetup[subfigure]{font=small,skip=2pt}
\begin{subfigure}[b]{\textwidth}
    \centering
    \includegraphics[width=\textwidth]{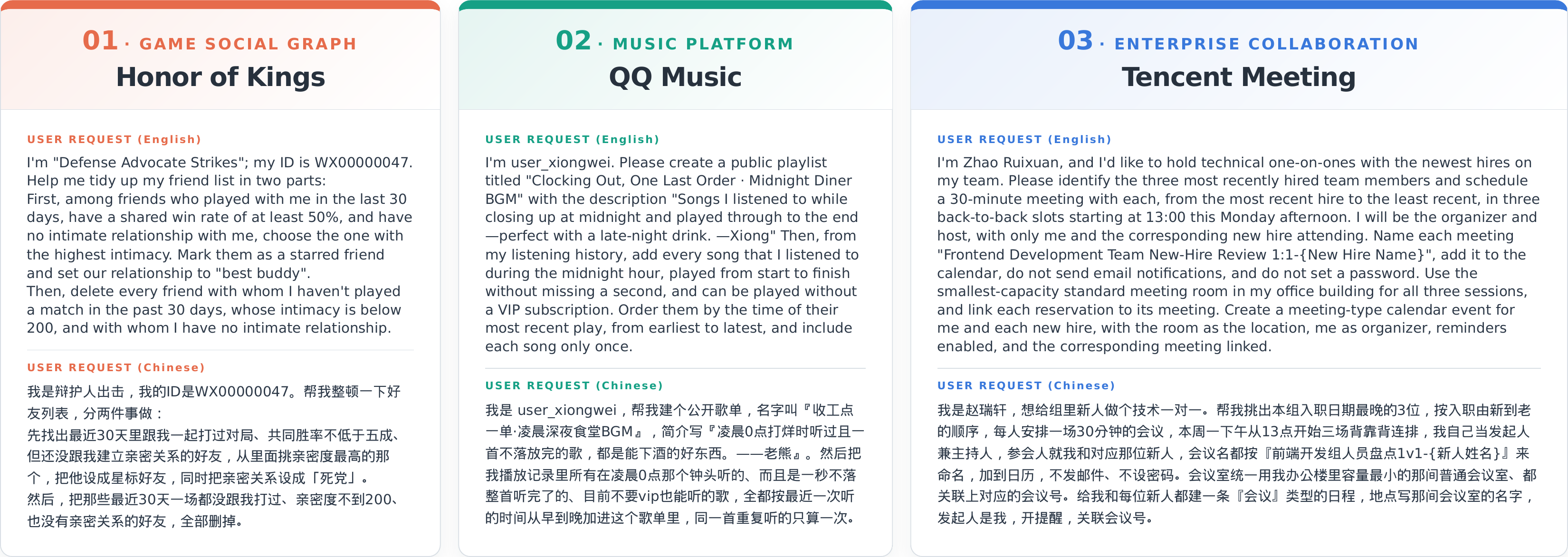}
    \caption{Representative natural-language tasks across the three domains.}
    \label{fig:overview-queries}
\end{subfigure}
\begin{subfigure}[b]{\textwidth}
    \centering
    \includegraphics[width=\textwidth]{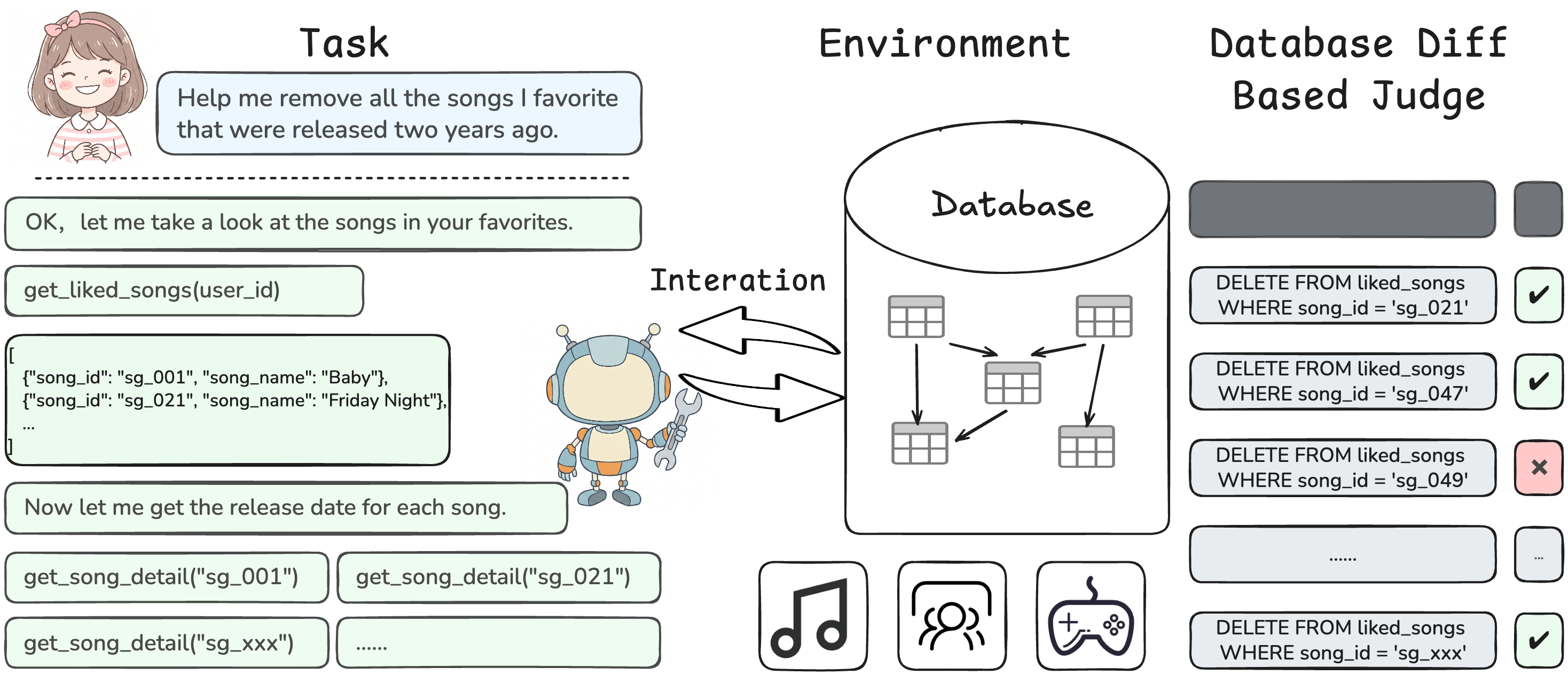}
    \caption{Tool-mediated task execution and deterministic database-diff evaluation.}
    \label{fig:overview-execution}
\end{subfigure}
\caption{Overview of \texttt{E-Bench}. 
(a) Natural-language tasks span diverse product domains. 
(b) Agents use domain-specific tools in fully-synthetic stateful environments, and are evaluated by comparing database-state diffs to ground truth.
}
\label{fig:overview_pipeline}
\vspace{-1em}
\end{figure}

This paradigm underlies many of the most valuable real-world applications, from operating software and querying databases to orchestrating business workflows, yet such environment-grounded agency remains hard to evaluate systematically. Recent benchmarks probe models' ability to invoke APIs correctly~\cite{patil2024gorilla, qin2024toolllm, patil2025the} and to operate in realistic web, OS, and software-engineering environments~\cite{zhou2024webarena, trivedi2024appworld, jimenez2024swebench}. While these reveal weaknesses in planning, tool grounding, and long-horizon execution, they center on short tool-use trajectories~\cite{yao2024tau, mialon2023gaia}, isolated API calls~\cite{patil2024gorilla, patil2025the}, or static QA~\cite{rein2024gpqa, hendrycks2021measuring, yang2018hotpotqa} without environment modification, and thus fail to capture agentic capability under partial observability, heterogeneous tools, long-horizon dependencies, and precise state-changing operations.

Moreover, although benchmarks built on real systems offer rich, stateful interactions, they are hard to scale, expensive to annotate, sensitive to data-safety and privacy constraints, and unstable to reproduce as the underlying services evolve~\cite{yao2024tau, liu2024agentbench, xu2024theagentcompany}. They can also be difficult to decontaminate: when tasks depend on public facts, familiar interfaces, or recurring workflows, model performance may reflect prior exposure rather than the ability to actively acquire hidden state through tools. 
A faithful benchmark must therefore present a complete, controllable environment in which the model cannot shortcut the intended reasoning, while remaining cheap to scale and safe to release.

To this end, we introduce \verb|E-Bench|, a \textbf{\textit{fully synthetic}} benchmark for evaluating multi-step tool-use agents. \verb|E-Bench| comprises $323$ state-changing tasks spanning three domains modeled after real products: \textit{Honor of Kings}, \textit{QQ Music}, and \textit{Tencent Meeting}. As illustrated in Figure~\ref{fig:overview_pipeline}, these tasks take the form of realistic natural-language requests (Figure~\ref{fig:overview-queries}) and are solved through tool-mediated interaction with a stateful product environment, with correctness determined by exact database-state differences (Figure~\ref{fig:overview-execution}).

\verb|E-Bench| is built in two decoupled stages: \emph{environment synthesis} and \emph{task synthesis}. For each domain, we first construct a reusable, fully mocked product environment rather than task-specific state snapshots, and then synthesize many state-changing tasks from this shared environment.
In the first stage, we synthesize a large, diverse, and consistent database via \emph{graph-guided database filling}.
Specifically, we construct a table-dependency graph from the relational schema and use a powerful LLM to populate tables in topological order, enforcing referential integrity by construction. 
This yields an orphan-free, stateful product world rather than tool stubs or task-local fixtures.
In the second stage, a privileged task \emph{generator} with SQL and code access explores this environment and authors tasks through a realistic product loop: \emph{inspect data $\rightarrow$ decide targets $\rightarrow$ modify state $\rightarrow$ summarize intent}. For each task, it records the induced database change as a ground-truth diff and annotates the exercised tool-use capabilities. 
Since tasks are derived from a shared reusable environment, new tasks can be generated by varying target entities, constraints, and reasoning patterns without redesigning the underlying data. Moreover, because all tasks share a complete underlying environment rather than task-specific, incomplete local snapshots, solvers have greater room to explore the available context, better reflecting the contextual richness encountered in real-world scenarios. 

During benchmarking, the agent (the \emph{solver}) has no direct database or SQL access. In the base \verb|E-Bench| setting, it is also denied code execution, while \verb|E-Bench|-Code grants this capability. This generator-solver asymmetry creates an \textit{information gap} and a \textit{tool gap}: solvers must discover hidden data, reason from partial observations, and compose available tools (e.g., issuing parallel independent calls) to produce the correct state change. Outcomes are graded deterministically by comparing the final database state with the ground-truth diff, without any LLM judge.

By evaluating agents through multi-step tool interaction with simulated product environments, \verb|E-Bench| measures their ability to gather necessary information, coordinate tool calls, and decide what to retrieve before acting. Since both environments and tasks are synthetically constructed rather than drawn from real users or live services, \verb|E-Bench| is controllable at the environment level and scalable at the task level for training-oriented use.
Our contributions are as follows:

$(1)$ We introduce \verb|E-Bench|, a fully synthetic benchmark for systematic evaluation of LLM agents on \emph{multi-step tool use}, which comprises $323$ state-changing tasks across three domains modeled after real corporate products.

$(2)$ We design a fully synthetic construction pipeline that decouples \emph{environment synthesis} from \emph{task synthesis}, enabling a single reusable, integrity-preserving environment to support scalable task generation. 
Evaluation uses deterministic database-state diffs rather than LLM judges.

$(3)$ We benchmark $11$ cutting-edge LLMs as tool-use agents and find that multi-step tool use remains far from solved: the strongest model (Kimi-K3) reaches only $73.8\%$ Avg@3, while consistency-oriented reliability, measured by Pass$^3$, remains below $60\%$. Granting code execution in \verb|E-Bench|-Code improves accuracy,
but performance consistency remains limited with Pass$^3$ below $70\%$.

\section{Related Work}

\textbf{Tool-Use and Function-Calling Benchmarks.}
Early tool-use benchmarks evaluate whether models can select appropriate tools and generate well-formed calls. Gorilla~\cite{patil2024gorilla} connects LLMs to large API collections, ToolLLM~\cite{qin2024toolllm} scales instruction-following data to thousands of real-world APIs, and the Berkeley Function Calling Leaderboard (BFCL)~\cite{patil2025the} standardizes function-calling evaluation across single-call, parallel-call, multi-turn, and multi-step settings. These benchmarks are valuable for measuring API grounding and call-structure correctness, but they primarily view tool use as producing correct invocations. In contrast, \verb|E-Bench| evaluates closed-loop interaction with a persistent environment, where agents must acquire information from observations, reason across steps, and execute actions whose effects modify backend product state.

\textbf{Stateful Agent Environments for Tool Use Benchmarks.} 
A second line of work evaluates agents in richer, stateful environments, but often compromises controllability, reusability, or automated construction. Benchmarks built on real or live MCP servers, such as MCP-Atlas~\cite{bandi2026mcpatlas}, MCP-Universe~\cite{luo2025mcpuniverse}, MCP-Bench~\cite{wang2025mcpbench}, and LiveMCPBench~\cite{mo2025livemcpbench}, provide high ecological validity but are difficult to reset, scale, and use for repeated state-changing evaluation. Controlled simulators avoid live services, yet often couple environments to individual tasks, as in AppWorld~\cite{trivedi2024appworld} and VitaBench~\cite{he2025vitabench}. More broadly, benchmarks such as VitaBench~\cite{he2025vitabench}, $\tau$-bench~\cite{yao2024tau}, ToolSandbox~\cite{lu-etal-2025-toolsandbox}, and MCPMark~\cite{wu2025mcpmark} require substantial human or engineering effort to package user requests, annotate tasks, create ground truth, or build custom checkers respectively. MCPEval~\cite{liu2025mcpeval} moves toward automation by generating tasks from MCP tool specifications and verifying them through frontier-agent execution; however, it evaluates agents by alignment to verifier-generated tool-use trajectories, which can favor a particular solution path over the final outcome.

\verb|E-Bench| addresses these limitations by decoupling automatic environment and task construction: each domain is a reusable, controllable, database-backed product world shared across many closed-loop tasks. LLM agents then synthesize tasks by inspecting this environment, executing intended state changes, and recording verified database diffs as trajectory-agnostic ground truth, enabling deterministic grading without task-specific engineering or LLM-based evaluation.

\section{E-Bench: A Benchmark for Multi-Step Tool Use} \label{sec:e_bench}

\begin{figure}[!t]
    \centering
    \includegraphics[width=\textwidth]{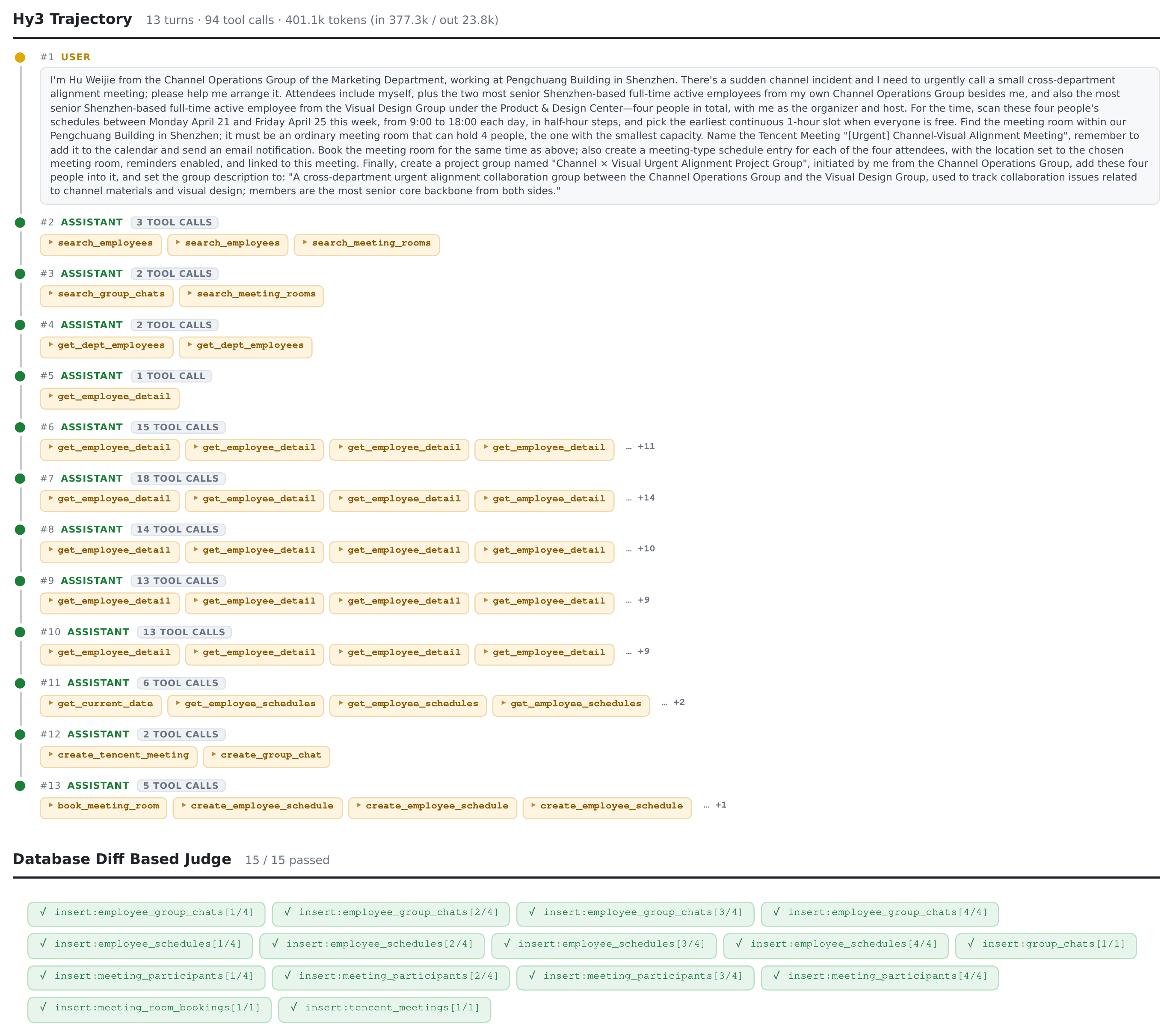}
    \caption{Hy3 trajectory on the Tencent Meeting case. The model successfully coordinates participant selection, schedule search, room reservation, meeting creation, calendar updates, and group creation.}
    \label{fig:trajectory-tencent-meeting}
\end{figure}

\subsection{Overview} \label{subsec:e_bench_overview}

\verb|E-Bench| follows one principle: evaluating multi-step tool use requires both a coherent environment and tasks that cannot be solved without interacting with it. To this end, \verb|E-Bench| explicitly decouples \emph{environment synthesis} from \emph{task synthesis}: instead of constructing a task-specific state snapshot for each task, we first build a reusable, fully mocked product environment for each domain, and then synthesize many state-changing tasks from this shared environment state.

The construction of \verb|E-Bench| therefore has two stages. First, we construct a \textbf{\textit{purely synthetic product environment}} from a relational schema and populate it through \textbf{\textit{graph-guided database filling}}, yielding a coherent, orphan-free database-backed product world rather than isolated tool stubs or task-local fixtures. Second, we synthesize tasks on top of the pre-populated environment through a controlled \textbf{\textit{generator-solver asymmetry}}: the generator can inspect the full database and use privileged tools, whereas the solver observes the environment only through restricted domain-specific MCP tools. This introduces \textbf{\textit{an information gap and a tool gap}} by design, forcing solvers to recover hidden state, compose multi-step and often parallel tool calls, and execute verifiable state changes. Because tasks are generated from a reusable shared environment rather than hand-authored with task-specific data fixtures, task construction becomes scalable: new tasks can be derived by varying target entities, constraints, and reasoning patterns without redesigning the underlying environment data for each task. \verb|E-Bench| instantiates this pipeline across three domains---\emph{Honor of Kings}, \emph{QQ Music}, and \emph{Tencent Meeting}---producing $323$ state-changing tasks. We show one task example from \emph{Tencent Meeting} in Figure~\ref{fig:trajectory-tencent-meeting} with more included in the Appendix~\ref{sec:trajectory-cases}.

The remainder of this section is organized as follows. Section~\ref{subsec:env-construction} details graph-guided environment construction; Section~\ref{subsec:task-construction} describes automatic task construction and deterministic evaluation; Section~\ref{subsec:e_bench_cli} introduces \verb|E-Bench|-Code, which grants solvers code execution to partially close the tool gap; and Section~\ref{subsec:benchmark_statistic} presents the comprehensive statistical summarization of \verb|E-Bench|. Together, these designs enable \verb|E-Bench| to test \textbf{\textit{full-information acquisition and multi-step parallel tool use}} under realistic product-state feedback.

\subsection{Reliable Environment Construction} \label{subsec:env-construction}

\texttt{E-Bench} first constructs, for each domain, a product environment fully simulated by a database. Each domain is defined by a set of tables, where each table corresponds to a concrete entity or relation. For example, in the music domain, tables represent users, artists, albums, songs, playlists, playlist-song relations, and user favorites; in the meeting domain, they represent employees, departments, meeting rooms, meetings, participants, and employee schedules. Each table further contains fields describing entity attributes, such as an album's release date, a song's lyrics and copyright status, or a meeting's start time and room location. The goal of environment construction is therefore not to generate a small set of plausible examples, but to build a product-level data world with diverse entity types, multi-hop relations, and historical states.

\begin{figure}[t]
\centering
    \includegraphics[width=0.95\textwidth]{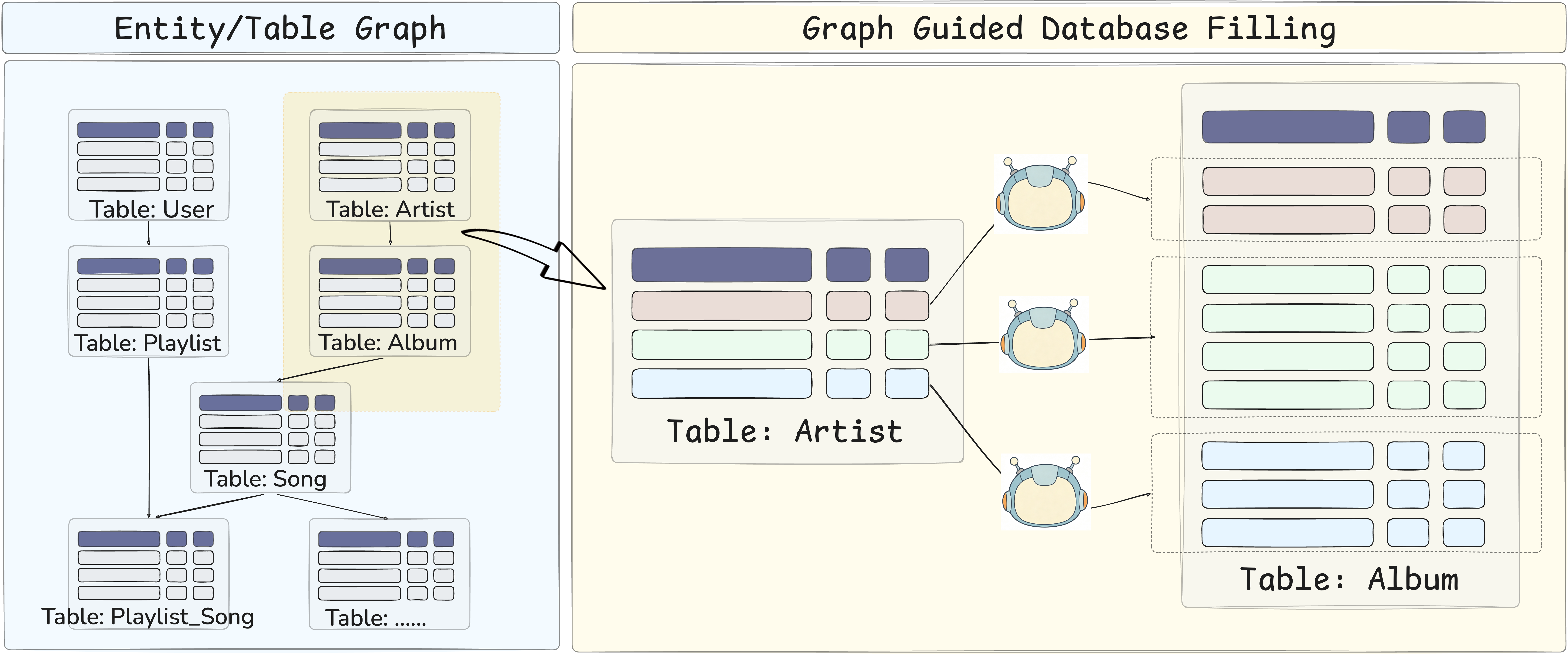}
\caption{Demonstration of graph-guided database filling and constrained context-aware synthesis.}
\label{fig:database_filling}
\end{figure}

\textbf{Graph-Guided Database Filling.}
Tables within each domain are linked by explicit foreign-key dependencies: for example, albums reference existing artists, songs reference existing albums, and playlist-song relations reference both existing playlists and songs. These dependencies determine record validity, since a row is meaningful only when its referenced entities exist and are semantically consistent. \texttt{E-Bench} therefore converts each relational schema into a table-level dependency graph (see also the left panel in Figure~\ref{fig:database_filling}), initializes an empty database with primary-key and foreign-key constraints, enables foreign-key checking, and populates tables in topological order. Root tables such as those for artists, departments, and meeting rooms are generated first, followed by downstream tables only after their dependencies are materialized. This process guarantees referential integrity by construction rather than repairing violations after free-form generation.

\textbf{Constrained Context-Aware Synthesis.}
During database synthesis, \texttt{E-Bench} uses a powerful LLM as a constrained synthesizer rather than asking it to generate an entire database freely. When populating a table, the model receives the table schema, key annotations, generation requirements, global domain context, and relevant existing records. For tables dependent on upstream entities, it generates child records conditioned on each parent, like shown in Figure~\ref{fig:database_filling}. For example, when creating playlist-song relations, the model observes the current playlist and sampled candidate songs, and may issue read-only queries if additional candidates are needed.
This process is constrained by tool-mediated interaction with the database. The model inserts records only through insertion tools, allowing the database to enforce primary-key uniqueness and foreign-key validity at insertion time. For downstream tables that must select from existing objects, the model is forced to query candidate entities from the current database state rather than inventing identifiers. The system also samples relevant records and retrieves parent-related context along the dependency graph, keeping prompts compact while preserving cross-table consistency.

\textbf{Post-Generation Validation and Repair.} 
After database population, \texttt{E-Bench} applies deterministic validation and repair scripts to ensure the fidelity and integrity of the constructed environment. These scripts correct semantic errors that may remain in LLM-synthesized data, such as inconsistent temporal ordering, incorrect aggregate counts, or mismatched status fields, by verifying and automatically repairing cross-field, cross-table, and temporal constraints, improving the consistency and quality of the final environment.

\textbf{Domain-Specific Tool Construction.}
After validation, \texttt{E-Bench} builds domain-specific MCP tools on top of the database. These tools function as CRUD interfaces over the underlying database, but expose product-level semantics to agents, such as searching songs, viewing playlists, adding favorites, querying meeting rooms, creating meetings, and updating schedules. Because all tools operate over the same pre-populated, referentially complete database, agents receive realistic product feedback during interaction: a retrieved song is linked to its album and artist, a meeting is linked to its participants and room, and a playlist is linked to songs that actually exist. We further validate the implementations of these MCP tools with unit tests to ensure that tool behavior remains consistent with database semantics.

Overall, our carefully designed construction pipeline avoids the incompleteness common in task-coupled benchmarks, in which the benchmark environment may contain merely the objects needed for a specific task. For example, an agent may retrieve an entity but fail to access its dependencies or related context, or encounter downstream records that reference missing upstream entities, i.e., orphan records. Through foreign-key constraints, graph-guided database filling, constrained context-aware synthesis, controlled insertion, as well as deterministic validation and repair, \texttt{E-Bench} ensures that dependency fields are resolved during record generation and that orphan records are ruled out by construction. The result is not merely a reliable fully synthesized database, but a structurally complete, cross-table consistent, and tool-interactive product state space that provides a stable foundation for automatic task synthesis.

\subsection{Scalable Task Construction} \label{subsec:task-construction}

\begin{table}[t]
\centering
\small
\setlength{\tabcolsep}{3pt}
\renewcommand{\arraystretch}{1.12}
\caption{
    Capability types annotated on \texttt{E-Bench} tasks. Each task may compose multiple capabilities.}
\label{tab:capability_dimensions}
\begin{tabularx}{\linewidth}{@{}l X rrrr@{}}
\toprule
\textbf{Capability} & \textbf{What the task requires} & \textbf{HoK} & \textbf{Music} & \textbf{Meeting} & \textbf{Total} \\
\midrule
Full-Data Acquisition & Retrieve complete hidden state before acting, avoiding decisions from partial observations. & 86 & 92 & 19 & 197 \\
Multi-Condition Filtering & Identify target entities satisfying multiple simultaneous constraints or states. & 36 & 20 & 16 & 72 \\
Aggregation and Computation & Compute counts, rankings, summaries, or other aggregates over retrieved state. & 56 & 66 & 37 & 159 \\
Cross-Step Dependency & Coordinate sequential tool calls where later steps depend on earlier results. & 78 & 65 & 64 & 207 \\
Precise Boundary Judgment & Make exact threshold, boundary, ranking, capacity, or availability decisions. & 35 & 15 & 94 & 144 \\
Cross-Entity Cascade & Propagate an operation or decision across related entity types. & 5 & 8 & 93 & 106 \\
\bottomrule
\end{tabularx}
\end{table}

\begin{figure}[t]
\centering
    \includegraphics[width=\textwidth]{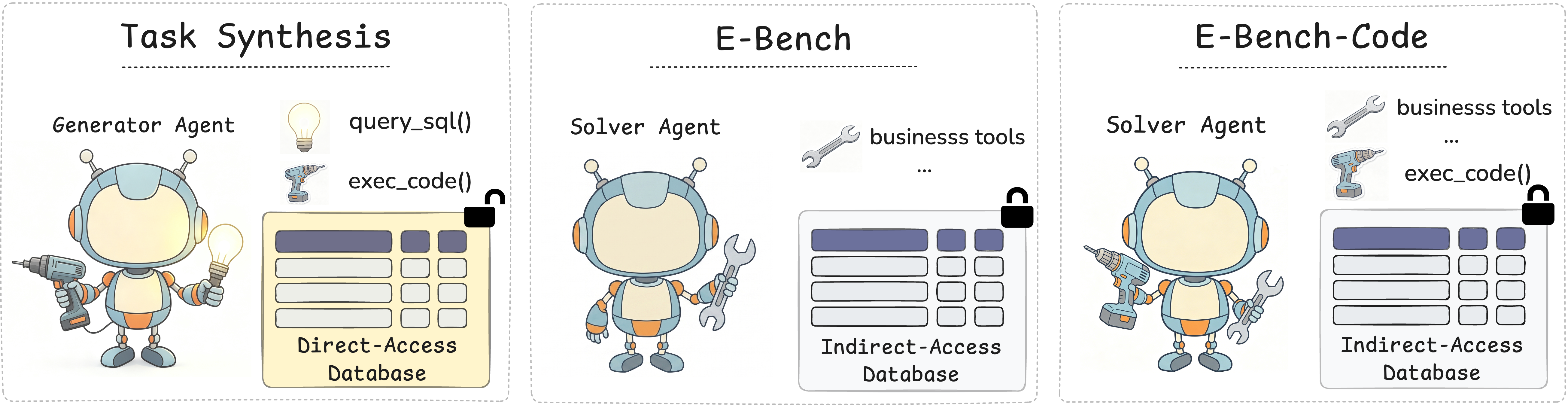}
\caption{Demonstration of the \emph{information gap} and \emph{tool gap} between \emph{generator} and \emph{solver} under \texttt{E-Bench} and \texttt{E-Bench}-Code respectively.}
\label{fig:information_tool_gap}
\end{figure}

The second stage turns each environment into benchmark tasks under one principle: a task is worth including only if solving it genuinely requires interacting with the environment.
\texttt{E-Bench} enforces this through a controlled asymmetry between task \emph{generation} and \emph{solving}: during generation, a strong \emph{task generator} (Claude Opus 4.7) works on a temporary database copy and may inspect global state and run code, whereas at evaluation the \emph{solver} sees only a natural-language request and the public domain tools.
This lets \verb|E-Bench| author tasks that are deterministically solvable yet cannot be shortcut, forcing agents to recover hidden information and induce the required state change through ordinary tool calls.
The asymmetry has two faces---an \underline{\textit{information gap}} and a \underline{\textit{tool gap}}, introduced by construction---which we describe next (see also Figure~\ref{fig:information_tool_gap}).

\textbf{The Information Gap.}
The information gap withholds global database state from the solver, so a task cannot be answered from its prompt alone.
Seeing the full schema and read-only \texttt{query\_sql}, the generator can define tasks whose targets depend on latent state rather than named values---e.g., all songs meeting a hidden condition, the intersection of two sets, the top entity under an aggregation, entities several foreign-key hops away, or the earliest feasible meeting slot.
This knowledge is never leaked: an intent-rewriting step keeps user-specified values explicit (a playlist name, a date range) but replaces database-derived values with the \emph{rule} producing them, so the query says ``add the unavailable songs in my favorites'' rather than listing song IDs or their count.
The gap thus targets \underline{\textit{full-information acquisition}}: it tests whether an agent can actively gather the \emph{complete} hidden target set before acting, rather than commit prematurely on partial evidence.

\textbf{The Tool Gap.}
The tool gap makes the generator computationally stronger than the base solver. During task generation, the strong task generator can invoke internal tools such as \texttt{query\_sql} and \texttt{exec\_code}: the former accesses the database directly, while the latter exposes public tools as callable functions, enabling programs that paginate results, compute aggregates, perform set operations, traverse multi-hop relations, and chain dependent writes. This enables the construction of tasks whose ground truth requires genuine computation rather than a single domain-specific tool call.
At evaluation time, internal tools are removed, leaving agents with only business-level domain-specific MCP tools; the solver must therefore reproduce the target outcome through ordinary tool calls. This gap tests whether agents can correctly compose and order multi-step tool use---including issuing independent \underline{\textit{parallel tool calls}} when permitted by the environment---to perform computation they cannot directly offload. \verb|E-Bench|-Code partially closes this gap by granting solvers code execution, and comparing the two settings quantifies how much code execution raises the performance ceiling.

\textbf{Capability Taxonomy as Synthesis Guidance.} With the two gaps, we further define six capability types for generated tasks as listed in Table~\ref{tab:capability_dimensions}. These capabilities arise naturally from our benchmark design and fall into two broad groups. The first group, namely \emph{Full-Data Acquisition}, \emph{Multi-Condition Filtering}, and \emph{Aggregation and Computation}, is driven primarily by the information gap: agents must retrieve, aggregate, and compute over large volumes of hidden data. The tool gap further amplifies the difficulty of these tasks, but closing it can partially alleviate the burden. The second group, \emph{Cross-Step Dependency}, \emph{Precise Boundary Judgment}, and \emph{Cross-Entity Cascade}, on the other hand, places greater emphasis on reasoning and decision making, requiring agents to determine the next action from the information accumulated so far.

\textbf{Task Generation and Validation.}
Each task is generated through a product-oriented loop---\emph{inspect data $\rightarrow$ decide targets $\rightarrow$ modify state $\rightarrow$ summarize intent}---for a sampled user and capability type(s). We snapshot the database before and after the generator's writes, record the exact state diff as ground truth, and discard candidates with no state change, unparseable intent, or excessive tool use. Surviving tasks are then validated by three strong \emph{validators}---GPT-5.5, Claude Opus 4.7, and GLM-5.1---given the same full-information tools as the generator, \texttt{query\_sql} and \texttt{exec\_code}. This focuses validation on the correctness of the recorded state change rather than hidden-information recovery. We accept a task only if at least two validators reproduce changes consistent with the ground truth, and further filter out tasks solved easily by a weak baseline. The remaining tasks are stratified by capability type to keep a balanced mix of reasoning patterns.

Finally, each accepted task is stored as a self-contained JSON specification containing the user query, expected database changes, and exercised capability types. Since correctness is checked against deterministic database-state changes rather than an LLM judge, evaluation is stable and less sensitive to semantic-judging variance.

\subsection{\texttt{E-Bench}-Code: Closing the Tool Gap} \label{subsec:e_bench_cli}

Because \verb|E-Bench| withholds the code-execution capability from the solver, a \emph{tool gap} remains between the strong task generator and the solver (Section~\ref{subsec:task-construction}).
Our extension \verb|E-Bench|-Code partially closes it by granting the solver the generator's \verb|exec_code| tool, with which the agent writes Python that calls the domain-specific MCP tools as inner-functions---paginating, aggregating, and performing set operations programmatically instead of manually orchestrating long call sequences.
All else is held fixed (tasks, tools, database copies, and the state-diff evaluator), so any performance difference is attributable to this interface.

Importantly, \verb|E-Bench|-Code closes the \emph{tool gap} but not the \emph{information gap}: \verb|exec_code| exposes only the business-level domain-specific APIs, not \verb|query_sql| or raw database access, so agents must still discover hidden targets through the observable environment.
The two settings thus isolate complementary abilities: \verb|E-Bench| tests manual coordination over extremely long multi-step tool-call sequences, whereas \verb|E-Bench|-Code tests whether agents extract the relevant information. Comparing them reveals how much difficulty stems from orchestration mechanics versus reasoning about what to retrieve.

\subsection{Benchmark Statistics} \label{subsec:benchmark_statistic}

\begin{table}[t]
\centering
\small
\setlength{\tabcolsep}{6pt}
\renewcommand{\arraystretch}{1.15}
\caption{
Per-domain statistics of \texttt{E-Bench}. 
``Principal Entities'' stands for ``users'' for \emph{Honor of Kings} and \emph{QQ Music}, and ``employees'' for \emph{Tencent Meeting}; ``DB Diff'' stands for the ground-truth database changes; and ``Solver Tools'' stands for MCP tools available to the \emph{solver}, where \texttt{E-Bench}-Code additionally grants the \texttt{exec\_code} tool (one more MCP tool per domain).
Note that there is not DB Diff of update type for QQ Music by environment desgin.
}
\label{tab:benchmark_composition}
\resizebox{\textwidth}{!}{
    \begin{tabular}{lcccc}
    \toprule
        & \textbf{Honor of Kings} & \textbf{QQ Music} & \textbf{Tencent Meeting} & \textbf{Total} \\
    \midrule
        \# Tables & 16 & 12 & 13 & 41 \\
        \# Columns & 168 & 72 & 105 & 345 \\
        \# Columns / Table [min$\sim$max (mean)] & 4$\sim$22 (10.5) & 3$\sim$11 (6.0) & 2$\sim$22 (8.1) & - \\
        \# Foreign-key Relations & 29 & 16 & 24 & 69 \\
        \# Rows & 18{,}646 & 28{,}321 & 29{,}350 & 76{,}317 \\
        \# Principal Entities & 170 & 54 & 995 & 1{,}219 \\
    \midrule
        \# Tasks & 110 & 104 & 109 & 323 \\
    \midrule
        \# DB Diffs / Task [min$\sim$max (mean, median)] & 3$\sim$25 (9.5, 9) & 5$\sim$80 (16.5, 13) & 7$\sim$221 (48.5, 35) & - \\
        DB Diff Type [insert / update /delete] ($\%$) & 37.7 / 46.1 / 16.2 & 98.8 / 0.0 / 1.2 & 98.6 / 0.3 / 1.1 & - \\
    \midrule
        \# Solver Tools [\texttt{E-Bench} / \texttt{E-Bench}-Code] & 33 / 34 & 27 / 28 & 25 / 26 & - \\
        \bottomrule
    \end{tabular}
}
\end{table}

Table~\ref{tab:benchmark_composition} summarizes the scale and diversity of the constructed environments and tasks. Each domain forms a coherent product world rather than a thin collection of tool stubs: the synthetic databases contain $12$ to $16$ tables connected by $16$ to $29$ foreign-key relations, with $72$ to $168$ columns in total and $2$ to $22$ columns per table ($6.0$ to $10.5$ on average). Populating these schemas yields $18.6$K to $29.4$K rows per domain ($76.3$K total) and over $0.6$M data cells, organized around $170$ users (\emph{Honor of Kings}), $54$ users (\emph{QQ Music}), and $995$ employees (\emph{Tencent Meeting}) together with the heroes, songs, meetings, and other entities that reference them. Because tasks are synthesized from this shared, pre-populated state rather than task-local fixtures, each environment supports roughly one hundred distinct tasks. 
The design goals of each domain are listed below:
\begin{itemize}
    \item \emph{Honor of Kings}: A MOBA game platform covering player social interactions, team management, room-based matchmaking, and match review. Agents can query the current user's players, heroes, ranks, friends, teams, rooms, and match records, and perform actions such as adding friends, managing blacklists, creating or dissolving rooms, sending or responding to room invitations, approving team applications, favoriting matches, and purchasing heroes.
    \item \emph{QQ Music}: A music platform covering content search, user preferences, and playlist management. Agents can search and inspect songs, artists, albums, playlists, comments, and listening histories, and perform actions such as creating or deleting playlists, adding or removing songs, favoriting songs, and following artists for the current user.
    \item \emph{Tencent Meeting}: An enterprise collaboration platform covering organizational structure, group chats, meetings, meeting-room bookings, and employee schedules. Agents can query employees, departments, group chats, meetings, meeting rooms, and schedules, and perform actions such as creating meetings, booking meeting rooms, creating schedules, canceling meetings, managing group-chat members, and transferring employees across departments for the current employee.
\end{itemize}

The resulting tasks require substantive state changes rather than single-record edits. Each task is bound with a ground-truth database diff for grading, with an average of $24.9$ row-level changes across the benchmark, ranging from $9.5$ in \emph{Honor of Kings} to $48.5$ in \emph{Tencent Meeting}. Individual tasks induce between $3$ and $221$ changes. Change patterns are domain-dependent: \emph{Honor of Kings} tasks mix inserts, updates, and deletes, while \emph{QQ Music} and \emph{Tencent Meeting} are dominated by inserts, reflecting bulk operations such as adding favorites, booking meetings, and inviting participants. This diversity in scale and edit type requires agents to track many interdependent state changes rather than perform a single localized action. Moreover, as shown in Table~\ref{tab:capability_dimensions}, different domains emphasize different task categories, leading to differences in both difficulty and evaluation focus.

\section{Main Results}
\label{sec:main_results}

\begin{table*}[t] 
    \centering 
    \caption{Model performance on E-Bench and E-Bench-Code. Models are ranked by E-Bench Avg@3. 
    The highest three values in each column are highlighted in \textbf{bold}, \underline{underline} and \colorbox{gray!20}{gray background}.
    }\label{tab:ebench_cli_comparison}
    \begin{tabular}{c|l||ccc|ccc@{}}
        \toprule 
         & Model & \multicolumn{3}{c|}{\texttt{E-Bench}} & \multicolumn{3}{c}{\texttt{E-Bench}-Code} \\
         ~ & ~ & Avg@3 & Pass@3 & Pass$^3$ & Avg@3 & Pass@3 & Pass$^3$ \\
        \midrule
         1 & Kimi-K3 & \textbf{73.79\%} & \textbf{87.62\%} & \textbf{58.82\%} & \underline{77.61\%} & \underline{88.70\%} & \colorbox{gray!20}{65.80\%} \\
         2 & GPT-5.5 & \underline{72.03\%} & \colorbox{gray!20}{82.97\%} & \underline{57.59\%} & \colorbox{gray!20}{77.19\%} & \colorbox{gray!20}{87.32\%} & \underline{66.90\%} \\
         3 & Opus-4.8 & \colorbox{gray!20}{68.78\%} & \underline{84.33\%} & 50.81\% & \textbf{81.11\%} & \textbf{92.61\%} & \textbf{68.66\%} \\
         4 & Grok-4.5 & 66.10\% & 80.50\% & \colorbox{gray!20}{52.32\%} & 69.24\% & 82.66\% & 55.75\% \\
         5 & GLM-5.2 & 52.32\% & 71.52\% & 30.96\% & 60.99\% & 81.69\% & 42.25\% \\
         6 & Qwen-3.7-Max & 50.88\% & 70.59\% & 30.34\% & 61.92\% & 80.99\% & 44.01\% \\
         7 & Hy3 & 50.15\% & 74.30\% & 26.63\% & 64.40\% & 85.56\% & 44.01\% \\
         8 & Seed-2.1-Pro & 47.94\% & 65.33\% & 28.79\% & 53.04\% & 75.35\% & 30.99\% \\
         9 & Gemini-3.5-Flash & 42.62\% & 64.71\% & 21.98\% & 62.54\% & 82.39\% & 40.49\% \\
         10 & MiniMax-M3 & 41.07\% & 62.85\% & 20.12\% & 46.85\% & 74.65\% & 21.13\% \\
         11 & DeepSeek-V4-Pro & 34.47\% & 53.56\% & 17.34\% & 47.68\% & 72.89\% & 25.35\% \\
        \bottomrule 
    \end{tabular}
\end{table*}

\subsection{Evaluation Setup}

\textbf{Models.} We evaluate $11$ frontier LLMs, choosing a representative model from each developer at the time of our experiments: GPT-5.5, Opus-4.8, Grok-4.5, GLM-5.2, Qwen-3.7-Max, Hy3, Seed-2.1-Pro, Gemini-3.5-Flash, MiniMax-M3, Kimi-K3, DeepSeek-V4-Pro and Muse-Spark-1.1\footnote{Due to Meta's strict content moderation, Muse-Spark-1.1 results are reported only in Appendix~\ref{subsec:muse_spark_1.1}, Table~\ref{tab:muse_spark}.}. Each model is run as an agent with their highest available thinking effort using a shared harness which exposes each domain's MCP tools and manages the multi-turn interaction loop. 

\textbf{Experiment and metrics.} Each task is attempted in three independent trials, with every trial starting from a fresh, isolated copy of the domain database to prevent cross-trial interference. A trial succeeds only if the agent's final database state exactly matches the ground-truth diff under our deterministic rule-based verifier; otherwise, it receives no partial credit. We report three metrics of increasing strictness. \emph{Avg@3} is the per-trial success rate averaged over all tasks and trials, measuring typical performance on average. \emph{Pass@3} counts a task as solved if at least one of its three trials succeeds. \emph{Pass$^3$} requires all three trials to succeed, measuring reliability; the gap between Pass@3 and Pass$^3$ reveals how consistently a model reproduces the correct state change.

\textbf{Efficiency measures.} Beyond accuracy, we record the number of tool calls, number of agent action turns, and input/output tokens average across tasks
for evaluating interaction cost and characterizing agent efficiency.
We evaluate all models under both the base \texttt{E-Bench}, where solvers access only domain-specific MCP tools, and \texttt{E-Bench}-Code, which additionally provides \verb|exec_code|.

\subsection{Main Results}

\begin{figure}[t]
\centering
    \includegraphics[width=\textwidth]{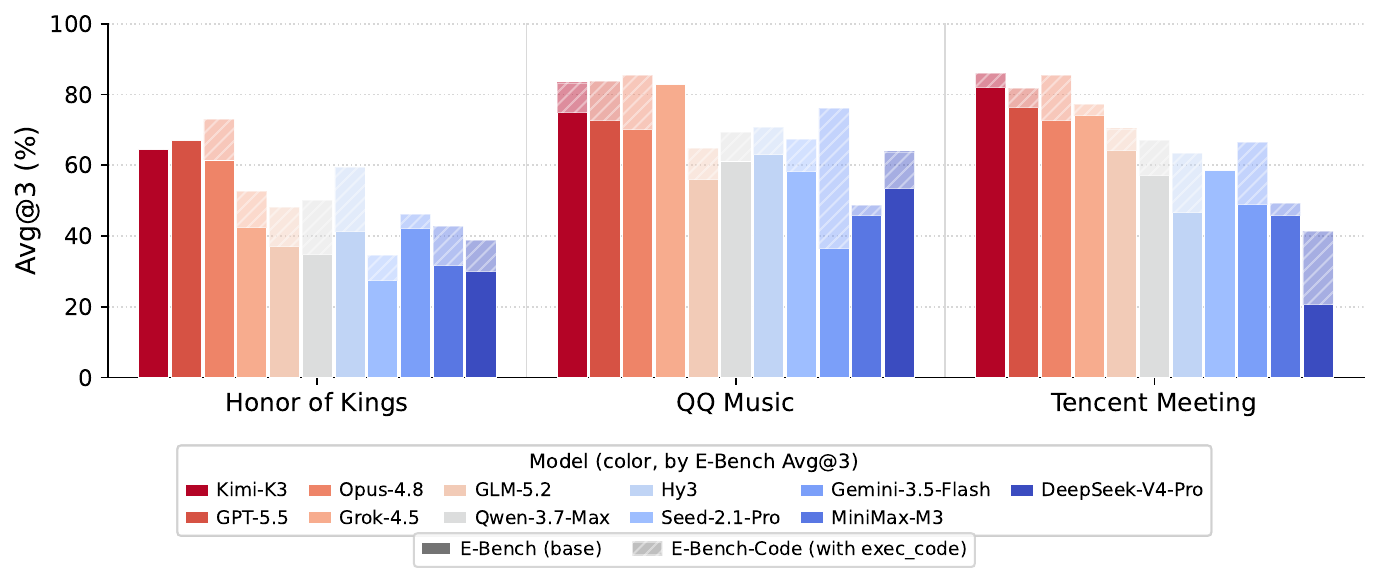}
    \caption{
    Per-domain Avg@3 (\%) for the $11$ evaluated models. 
    Models are ordered by \texttt{E-Bench} Avg@3.
    }
    \label{fig:ebench_cli_by_domain}
\end{figure}

\begin{figure}[t]
\centering
    \includegraphics[width=\textwidth]{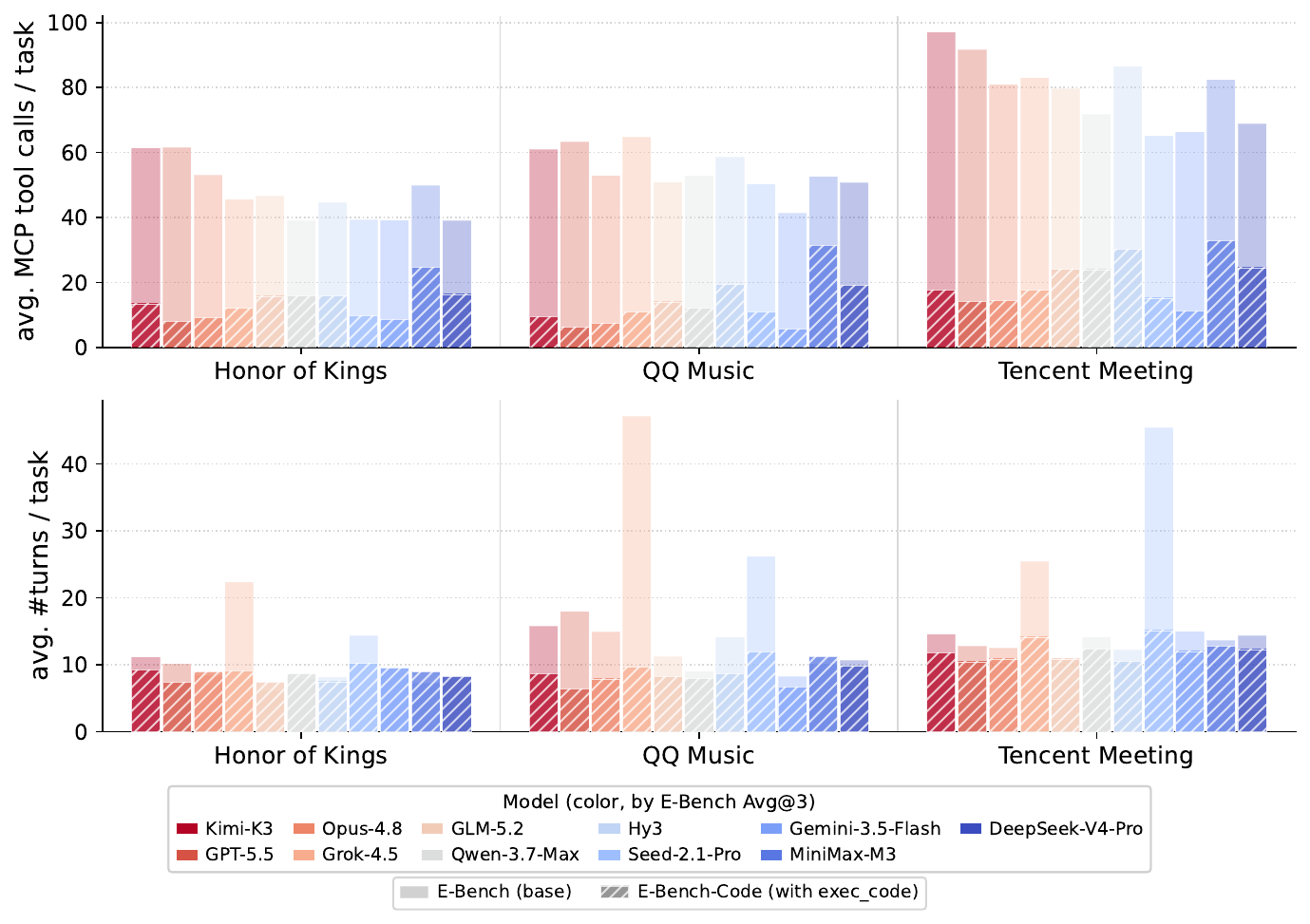}
    \caption{
    Per-domain MCP tool calls per task (top row) and number of turns per task (bottom row) for the $11$ evaluated models. 
    Models are ordered by \texttt{E-Bench} Avg@3.
    }
    \label{fig:ebench_cli_by_domain_toolcalls}
\end{figure}

Table~\ref{tab:ebench_cli_comparison} reports the performance of all evaluated models on \texttt{E-Bench}, ranked by Avg@3, together with \texttt{E-Bench}-Code results tested under the code-execution setting. 
Figure~\ref{fig:ebench_cli_by_domain} and Figure~\ref{fig:ebench_cli_by_domain_toolcalls} compare the per-domain Avg@3 and the per-domain interaction cost (number of MCP tool calls and agent action turns) for each model under these two settings.

\subsubsection{\texttt{E-Bench}}

\textbf{Multi-step tool use remains far from solved.}
Even the strongest agents leave substantial headroom on \texttt{E-Bench}. Kimi-K3 achieves the highest Avg@3 at $73.79\%$, followed by GPT-5.5, Opus-4.8, and Grok-4.5, all of which exceed $66\%$, while no other model exceeds $53\%$.
Also, across all $11$ models, Avg@3 averages only $54.56\%$. This indicates that a typical agent fails on nearly half of the tasks in a single attempt.

\textbf{Reliability remains the central bottleneck.}
The gap between Pass@3, Avg@3, and Pass$^3$ shows that many agents succeed only intermittently, with substantially higher Pass@3 compared to Pass$^3$. Even top models are non-robust: Kimi-K3 drops from $87.62\%$ Pass@3 to $58.82\%$ Pass$^3$, and GPT-5.5 drops from $82.97\%$ to $57.59\%$. For weaker models, reliability nearly collapses. As \texttt{E-Bench} requires exact state changes with no partial credit, such instability is practically significant: agents that solve tasks only occasionally are not yet dependable for modifying live product state.

\subsubsection{\texttt{E-Bench}-Code}
\textbf{Code execution raises the performance of all models, but reliability remains limited.}
Granting agents \texttt{exec\_code} in \texttt{E-Bench}-Code improves Avg@3 for all evaluated models, whether the model is strong or not. The gains are substantial: Opus-4.8 rises from $68.78\%$ to $81.11\%$, taking the top spot over Kimi-K3 and GPT-5.5; Gemini-3.5-Flash climbs from $42.62\%$ to $62.54\%$---a striking $46.7\%$ relative improvement; weaker models like DeepSeek-V4-Pro also gain a $38.32\%$ relative improvement.%
However, even with code execution, the best Pass$^3$ (Opus-4.8) remains below $70\%$ and the worst is only $21.13\%$, leaving substantial room for reliability improvement.

\textbf{Code execution reshuffles the leaderboard benefiting stronger code users.}
Models whose rankings rise on the leaderboard tend to be those that exploit \verb|exec_code| effectively: Gemini-3.5-Flash from $9^{th}$ to $6^{th}$, while Opus-4.8 (a known strong coder) takes the top position instead of the third place it holds in \texttt{E-Bench}. These models route a large fraction of their interaction through code (e.g., $95\%$ for Gemini-3.5-Flash and $93\%$ for Opus-4.8), replacing many individual tool calls with compact programs (i.e., often calling domain-specific functions inside \texttt{exec\_code}, rather than direct calling their corresponding MCP tools). We analyze this behavior further in Section~\ref{subsec:code_help_for_task_of_different_difficulty}.

\textbf{Code execution improves accuracy while cutting cost.}
The performance gains from \verb|exec_code| come with markedly \emph{lower} interaction cost. As shown in Figure~\ref{fig:ebench_cli_by_domain_toolcalls}, enabling \texttt{exec\_code} reduces the average number of MCP tool calls per task from $60.42$ to $15.86$ (a $73.8\%$ drop) and agent turns from $14.87$ to $9.87$ (a $33.6\%$ drop), averaged over $11$ models and three domains.
As stated in the previous section, the reduction comes from a single \verb|exec_code| block folding multi-turn domain-function calls into one top-level action.
Together with the Avg@3 gains in Figure~\ref{fig:ebench_cli_by_domain}, this shows that \texttt{exec\_code} enables models to solve more tasks while reaching solutions at lower cost. Further analysis of how these improvements are distributed across tasks appears in Section~\ref{subsec:code_help_for_task_of_different_difficulty}.

\subsection{How Does Difficulty Vary Across Domains? Per-domain Analysis}

We further analyze each domain separately. Figure~\ref{fig:ebench_cli_by_domain} plots per-model Avg@3 across the three domains.

\textbf{Domain difficulty and code-execution gains are both uneven.}
Figure~\ref{fig:ebench_cli_by_domain} shows that \emph{Honor of Kings} is the most challenging domain, while \emph{QQ Music} is the easiest. Under \texttt{E-Bench}, average Avg@3 is $43.64\%$, $58.87\%$, and $61.39\%$ for \emph{Honor of Kings}, \emph{Tencent Meeting}, and \emph{QQ Music}, respectively. With \texttt{E-Bench}-Code, these rise to $52.36\%$, $67.78\%$, and $71.94\%$. \emph{QQ Music} benefits most from code execution, gaining $10.55$ points on average, compared with $8.72$ and $8.91$ points for \emph{Honor of Kings} and \emph{Tencent Meeting}. This matches its capability mix: \emph{QQ Music} contains more tasks requiring retrieval and aggregation of large amounts of data (e.g. \emph{Full-Data Acquisition} and \emph{Aggregation and Computation} tasks), and fewer tasks requiring reasoning for decision making (e.g. \emph{Precise Boundary Judgment} tasks), which \verb|exec_code| can accelerate (Section~\ref{subsec:code_help_for_category}).

\textbf{Most model rankings vary across domains.}
Across domains and settings, Kimi-K3, GPT-5.5, and Opus-4.8 form the frontier tier, with Grok-4.5 as a strong second tier. However, rankings are highly domain-sensitive. For example, Grok-4.5 is the best base-\texttt{E-Bench} model on \emph{QQ Music} but only mid-pack on \emph{Honor of Kings}.
Below the frontier, models cluster closely and reorder substantially across domains.
Rankings also shift by setting, as \texttt{exec\_code} disproportionately benefits strong code users. 
These results show that robust multi-step tool-use evaluation requires coverage across diverse domains rather than focusing on a single domain.

\subsection{Where Does Code Execution Help? Per-Capability Analysis}\label{subsec:code_help_for_category}

\begin{table}[t]
\centering
\small
\setlength{\tabcolsep}{6pt}
\renewcommand{\arraystretch}{1.15}
\caption{
    Per-capability Avg@3 averaged over $11$ evaluated models under \texttt{E-Bench} and \texttt{E-Bench}-Code, as well as the absolute and relative gain of granting coding capability (``$\Delta$ Avg@3'' and ``Rel.\ $\Delta$ Avg@3''). Rows are sorted by gains ($\Delta$ Avg@3).
    The highest value in each column is highlighted in \textbf{bold}, and the second-highest is \underline{underlined}.
}
\label{tab:exec_code_lift}
\begin{tabular}{l||cc|cc}
\toprule
Capability & \texttt{E-Bench} & \texttt{E-Bench}-Code & $\Delta$ Avg@3 & Rel.\ $\Delta$ Avg@3 \\
\midrule
Multi-Condition Filtering   & 51.94\% & 62.55\% & \textbf{+10.61\%} & \textbf{+20.43\%} \\
Full-Data Acquisition       & 53.55\% & 63.65\% & \underline{+10.11\%} & \underline{+18.88\%} \\
Aggregation and Computation & 52.85\% & 62.30\% & +9.45\% & +17.88\% \\
Cross-Entity Cascade        & \textbf{60.08\%} & \textbf{68.88\%} & +8.80\% & +14.66\% \\
Cross-Step Dependency       & 54.11\% & 62.79\% & +8.68\% & +16.05\% \\
Precise Boundary Judgment   & \underline{57.39\%} & \underline{64.94\%} & +7.55\% & +13.16\% \\
\bottomrule
\end{tabular}
\end{table}

To further identify where \texttt{exec\_code} helps, Table~\ref{tab:exec_code_lift} decomposes the Avg@3 gain from adding \texttt{exec\_code} by capability, averaged over the $11$ evaluated models. 

\textbf{Code execution helps computation more than reasoning.}
As shown in Table~\ref{tab:exec_code_lift}, the largest gains from \texttt{exec\_code} appear in mechanically intensive capabilities: \emph{Multi-Condition Filtering}, \emph{Full-Data Acquisition}, and \emph{Aggregation and Computation}, 
with highest absolute and relative Avg@3 improvements at the same time.
These tasks require exhaustive retrieval, exact filtering, aggregation, and intermediate-result tracking, which code can handle reliably through loops and iterations. In contrast, gains are smaller for \emph{Cross-Step Dependency} and \emph{Precise Boundary Judgment},
as these capabilities depend more on selecting the right entities, dependencies, thresholds, or action order. Thus, \texttt{exec\_code} primarily offloads computation rather than reasoning and decision making, explaining both its large average gains and the remaining headroom in \texttt{E-Bench}-Code.

\subsection{When Do Models Choose to Code? Per-Task Analysis}\label{subsec:code_help_for_task_of_different_difficulty}

\begin{figure}[htbp]
\centering
    \includegraphics[width=\textwidth]{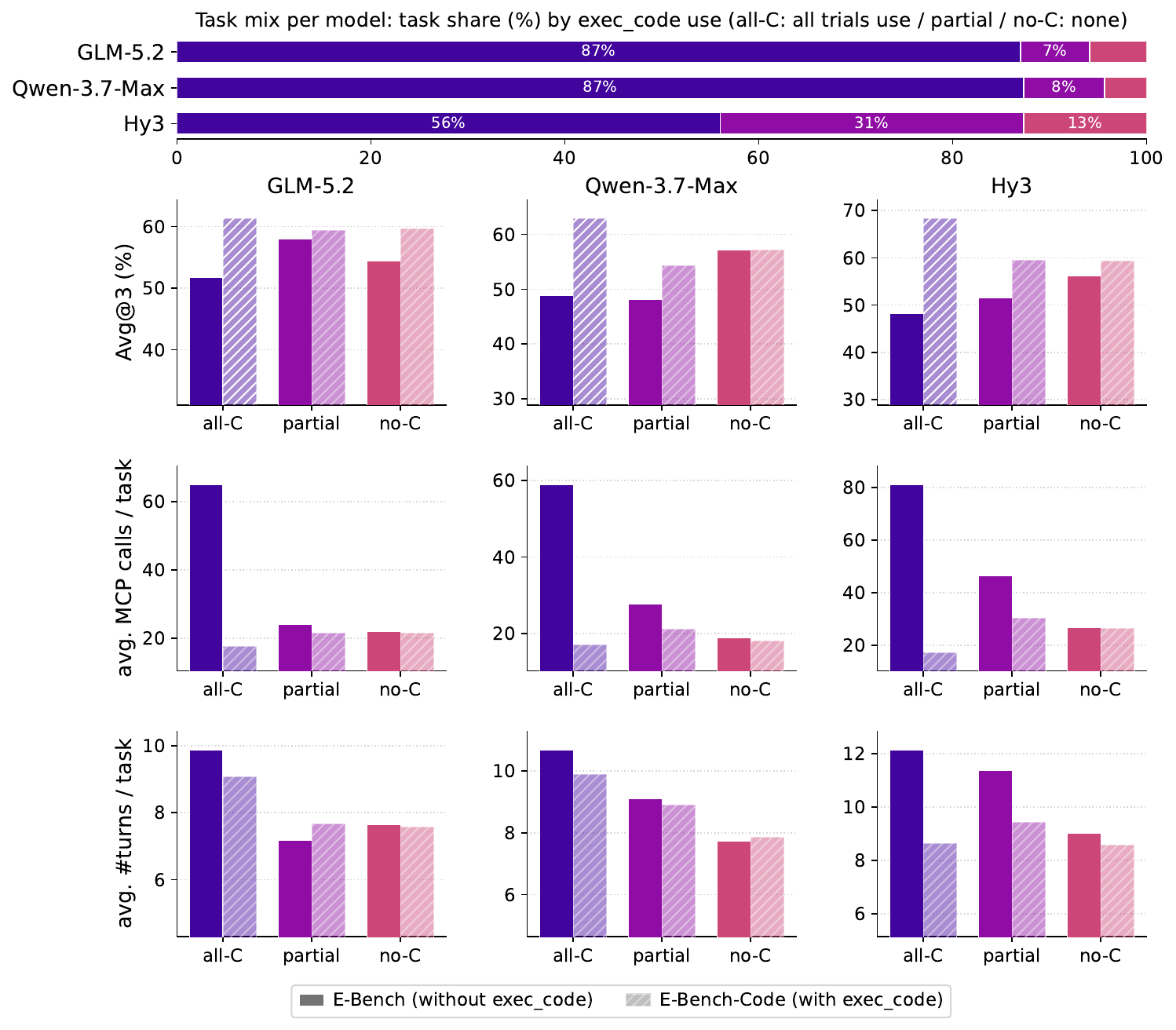}
    \caption{
        Comparison of task outcomes under \texttt{E-Bench}-Code, grouped by whether \emph{all} trials used \texttt{exec\_code} (all-C), \emph{none} used \texttt{exec\_code} (no-C), or usage was mixed across trials (partial). 
        Results are first averaged across trials and then across tasks.
        Results for three models are shown, with paired \texttt{E-Bench} results included as references.
    }
    \label{fig:exec_code_coverage}
\end{figure}

Another interesting observation is that, even when \texttt{exec\_code} is available under \texttt{E-Bench}-Code, many models still solve some of the tasks without using it, either consistently or sporadically across trials. Figure~\ref{fig:exec_code_coverage} groups tasks by \texttt{exec\_code} usage across three trials: all trials use it (all-C), no trials use it (no-C), or usage is mixed (partial). For these groups, we compare task-averaged performance (Avg@3), MCP tool calls, and agent action turns, with paired \texttt{E-Bench} results as references. For clarity, we show results for three representative models here and others in the Appendix~\ref{subsec:appendix_code_help_for_task_of_different_difficulty}.

\textbf{Code execution rescues hard tasks while reducing interaction cost.}
Tasks where models consistently avoid \texttt{exec\_code} (no-C) are relatively easy, often requiring the fewest MCP tool calls and shortest interaction turns. Accordingly, the two settings behave nearly identically on no-C tasks: tool calls and turns remain essentially unchanged, and Avg@3 fluctuates only slightly. In contrast, tasks where models consistently use \texttt{exec\_code} (all-C) are the most demanding, and \texttt{exec\_code} yields large performance gains while substantially reducing both MCP calls and agent turns, bringing costs close to those of no-C tasks. Mixed-usage tasks show the same trend: Avg@3 improves from $52.53\%$ to $57.70\%$, while MCP calls drop from $32.68$ to $24.34$ on average. Overall, \texttt{exec\_code} is used selectively on demanding, tool-heavy tasks, where it batches multi-step, often parallel, domain-specific tool calls into shorter execution traces controlled by more precise code.

This finding further explains the source of the substantial gains in performance and the reductions in interaction cost shown in Figure~\ref{fig:ebench_cli_by_domain} and Figure~\ref{fig:ebench_cli_by_domain_toolcalls}.

\subsection{Does Spending More Help? Performance versus Interaction Cost}
\label{subsec:pass1_vs_cost}

A natural question is whether agents that fail simply need to \emph{try harder}---issue more tool calls or consume more context.
Figure~\ref{fig:pass1_vs_cost} in the Appendix~\ref{subsec:apendix_pass1_vs_cost} plots Avg@3 against three per-task cost measures (i.e., task-average number of MCP tool calls, number of turns, and total tokens), with one point per <model, domain>.

\textbf{Harder tasks require more interaction.}
Under \texttt{E-Bench}, each action is an ordinary domain-tool call, and Avg@3 is moderately positively correlated with the number of tool calls across the $33$ <model,domain> pairs (Pearson $r=+0.57$). The correlation is even stronger within some domains ($r=+0.86$, $+0.61$, and $+0.89$ for \emph{Honor of Kings}, \emph{Tencent Meeting}, and \emph{QQ Music}, respectively). This matches the benchmark's intended \emph{information gap}: solvers must gather hidden information through multi-step, often parallel, tool use before committing state-changing actions. In this base setting, a low call count often indicates premature action rather than efficiency. Turns and total token consumption are also weakly positively correlated with success ($r=+0.39$ and $r=+0.38$), consistent with harder tasks requiring more interaction and deliberation.

In \texttt{E-Bench}-Code, by contrast, Avg@3 is weakly negatively correlated with the number of LLM-emitted tool calls across the same $33$ pairs (Pearson $r=-0.36$), with a much stronger negative correlation in \emph{QQ Music} ($r=-0.87$) and weaker ones in \emph{Tencent Meeting} and \emph{Honor of Kings} ($r=-0.57$ and $r=-0.36$). Because models can capability many domain-specific function calls into a single \texttt{exec\_code} block, fewer emitted tool calls do not imply less work. Rather, the result suggests that with code execution, it is possible to lower costs while preserving or improving performance (Avg@3). Additional analysis is provided in Appendix~\ref{subsec:apendix_pass1_vs_cost}.

More detailed analyses of efficiency and API cost (Figure~\ref{fig:price_vs_performance}) are provided in Appendix~\ref{sec:appendix_results}.
\section{Conclusion and Future Work}

In this work, we introduce \verb|E-Bench|, and its code-enabled extension \verb|E-Bench|-Code, a fully synthetic benchmark for evaluating LLM agents on \emph{multi-step tool use}. \verb|E-Bench| separates environment synthesis from task synthesis: graph-guided database filling builds reusable, orphan-free environments, while generator-solver asymmetry creates state-changing tasks with information and tool gaps. Agents must discover hidden data and compose multiple, often parallel, tool calls before changing state. Since both environments and tasks are synthetic, \verb|E-Bench| is controllable, scalable, and deterministically graded via database-state diffs.

Benchmarking $11$ frontier LLMs shows that multi-step tool use remains unsolved. The best model reaches only $73.79\%$ Avg@3, and Pass$^3$ remains below $60\%$ on \verb|E-Bench|. \verb|exec_code| within \verb|E-Bench|-Code improves every model, but even the best Pass$^3$ stays below $70\%$. The gains come from folding work into code: strong agents batch tool calls for computation-heavy data acquisition, aggregation, and filtering, while reasoning-heavy cross-entity cascades and boundary-sensitive decisions remain difficult.

Future work will extend \verb|E-Bench| toward real domain CLIs and live product back-ends, and from single-domain tasks to cross-domain scenarios that better reflect real product workflows. We also view \verb|E-Bench| as a controllable, scalable source of training data for improving agents' multi-step tool-use ability and reliability.

\bibliography{ref}

\newpage
\appendix
\section*{Appendix}

\section{More Benchmark Results Analysis}\label{sec:appendix_results}

\subsection{Parallel Tool Calling and Token Consumption}
\label{sec:parallel-tool-call}

Parallel tool calling is an important capability for agents operating in stateful environments, and it varies substantially across models. Many tasks in \texttt{E-Bench} require agents to explore multiple independent parts of the environment, such as retrieving details for several songs or checking the daily schedules of different users. In these cases, issuing independent tool calls in parallel can substantially improve efficiency by shortening interaction trajectories and reducing repeated context accumulation, thereby lowering token consumption.

We further analyze whether evaluated models can issue multiple tool calls within the same interaction round. For each model, we compute the average number of tool calls per turn as a proxy for parallel tool-calling capability, and compare it with the average token consumption per task. Our token-consumption statistic sums the input tokens sent to the model and the output tokens generated by the model across all model invocations in a trajectory.

\begin{figure}[htbp]
    \centering
    \includegraphics[width=0.9\textwidth]{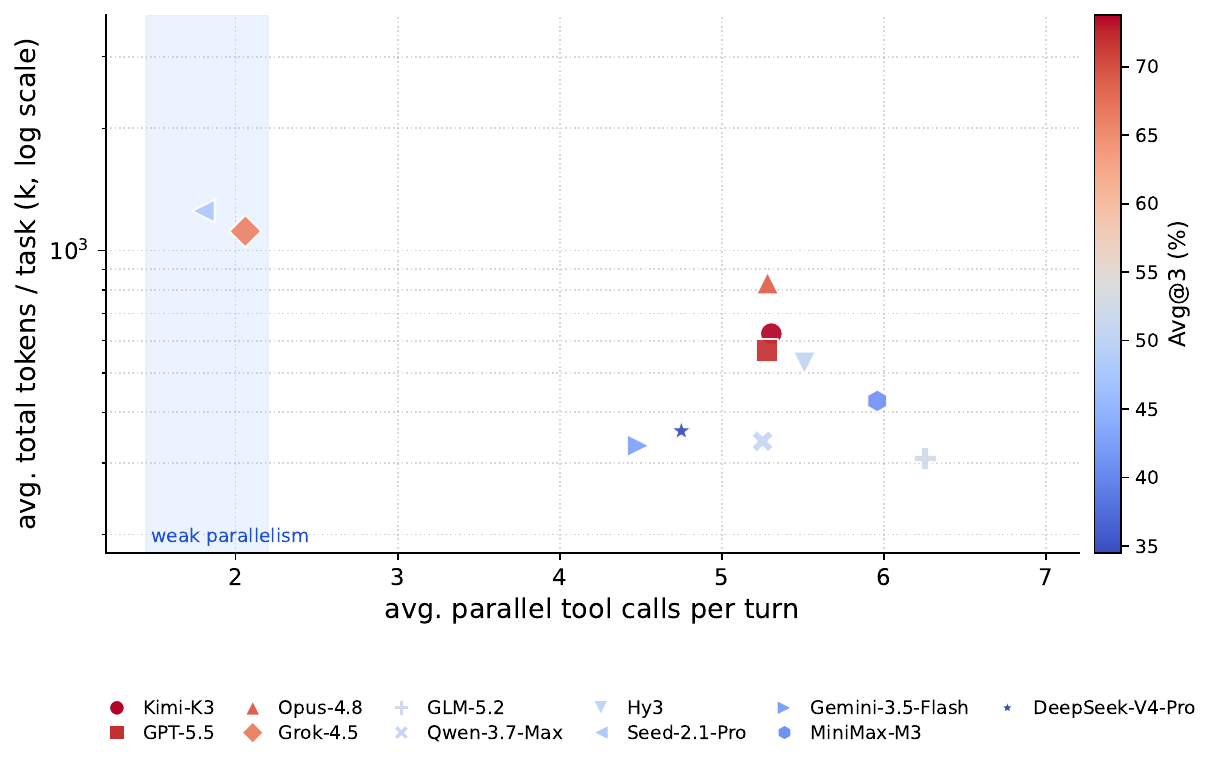}
    \caption{Relationship between global parallel tool-calling capability and token consumption. Each point denotes one evaluated model; marker shape identifies the model, and color indicates Avg@3. The x-axis measures the average number of tool calls issued per turn, and the y-axis shows the average total tokens consumed per task. }
    \label{fig:parallel-tool-calls-vs-tokens}
\end{figure}

As shown in Figure~\ref{fig:parallel-tool-calls-vs-tokens}, models that issue more MCP tool calls in each turn, for example GLM-5.2, usually finish the task with fewer interaction rounds, thereby avoiding repeatedly re-reading long observations and accumulating excessive context tokens. In contrast, Grok-4.5 and Seed-2.1-Pro exhibit notably weak parallelism. Grok-4.5 issues only 2.06 tool calls per turn on average and consumes 1.12M tokens per task, while Seed-2.1-Pro issues only 1.80 tool calls per turn and consumes 1.25M tokens per task. These two models consume 1.18M tokens per task on average, substantially higher than the 0.44M average of models with at least four tool calls per turn. This suggests that insufficient parallel tool invocation is an important source of token inefficiency in multi-step tool-use agents: weaker parallel tool-calling capability directly increases token usage and therefore leads to higher monetary cost under real API pricing.

\subsection{Per-Model Effect of Code Execution}
\label{subsec:appendix_code_help_for_task_of_different_difficulty}

\begin{figure}[htbp]
    \centering
    \includegraphics[width=\textwidth]{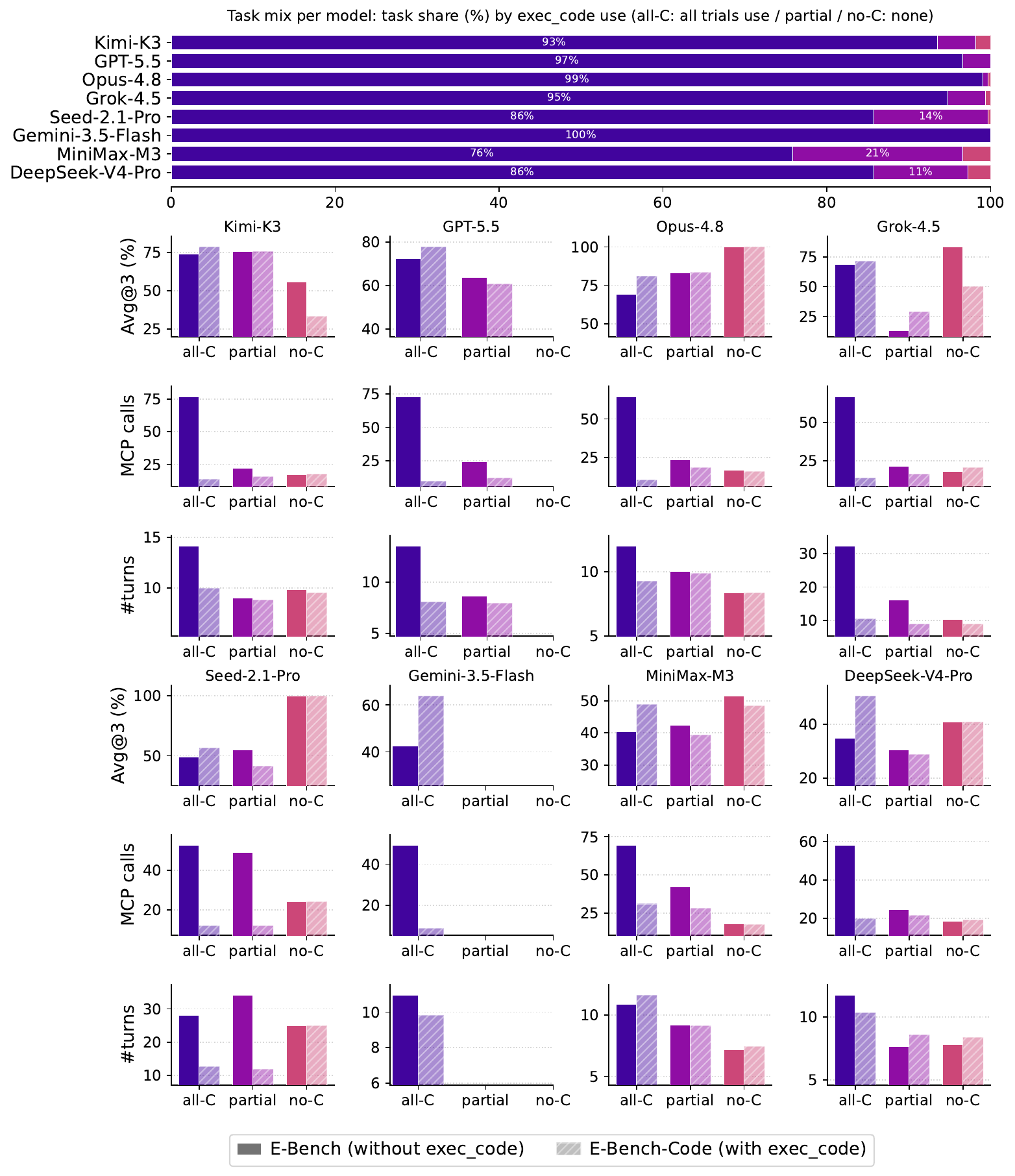}
    \caption{
    Effect of \texttt{exec\_code} for the eight models not shown in Figure~\ref{fig:exec_code_coverage}. 
    Bars are absent where a model has no tasks in a group (e.g. Gemini-3.5-Flash used \texttt{exec\_code} on every task, so it has only the all-C group).
    }
    \label{fig:exec_code_coverage_appendix}
\end{figure}

Section~\ref{subsec:code_help_for_task_of_different_difficulty} analyzed how \verb|exec_code| affects performance and interaction cost for the three models with a paired \texttt{E-Bench} run shown in the main text (GLM-5.2, Qwen-3.7-Max, Hy3). Here we report the same metrics on different kinds of task for the remaining eight models, in Figure~\ref{fig:exec_code_coverage_appendix}.

The pattern in Figure~\ref{fig:exec_code_coverage_appendix} mirrors that of Figure~\ref{fig:exec_code_coverage} in the main text.

First, tasks for which a model always uses code (all-C) are the most tool-intensive. Under \verb|E-Bench|, where \texttt{exec\_code} is unavailable, models must issue many individual domain-specific MCP calls to solve these tasks, leading to substantially higher call counts compared to the other two types of tasks. Granting \texttt{exec\_code} sharply reduces the number of MCP calls; Avg@3 often improves as well, and the number of turns often decreases. This confirms that a single code block can replace a long sequence of individual tool calls without sacrificing accuracy.

A caveat is that some models rely on \texttt{exec\_code} so heavily that their \emph{partial} and \emph{no-C} groups contain very few tasks: Gemini-3.5-Flash has no tasks outside all-C; GPT-5.5 has no no-C tasks; Opus-4.8 assigns $99\%$ of tasks to all-C; Grok-4.5 assigns $95\%$ of tasks to all-C with only two no-C tasks out of 323 tasks in total; and Kimi-K3 likewise assigns $94\%$ of tasks to all-C with only six no-C tasks. Thus, their no-C and partial bars should be interpreted as indicative rather than statistically robust. The trends are clearest for models with sizeable groups in all three categories (Seed-2.1-Pro, MiniMax-M3, and DeepSeek-V4-Pro): for these models, no-C tasks show nearly identical cost and performance across the two settings, as expected when code is not used.

\subsection{Does Spending More Help in \texttt{E-Bench}-Code?}
\label{subsec:apendix_pass1_vs_cost}

\begin{figure}[htbp]
\centering
    \includegraphics[width=\textwidth]{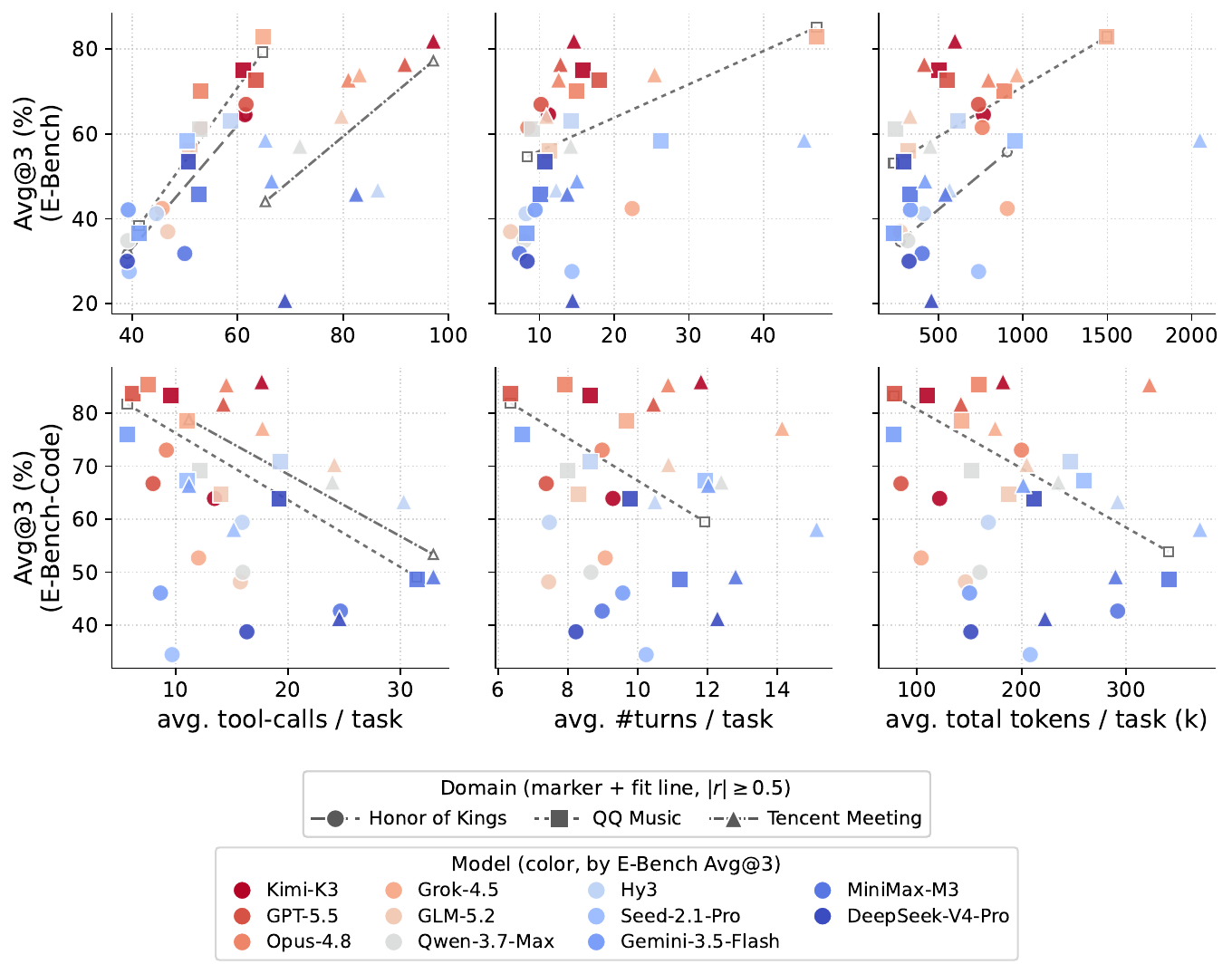}
    \caption{
    Avg@3 (\%) versus per-task interaction cost (task-average MCP tool calls, number of turns, and total tokens) for all evaluated models in three domains under \texttt{E-Bench} (top row) and \texttt{E-Bench}-Code (bottom row). Gray least-squares lines, matching the domain markers in the legend are drawn in any panel where the within-domain Pearson correlation satisfies $|r|\ge0.5$.
    }
    \label{fig:pass1_vs_cost}
\end{figure}

As discussed in Section~\ref{subsec:apendix_pass1_vs_cost}, under \verb|E-Bench|, higher interaction cost is associated with better performance (Avg@3) on harder tasks: stronger runs tend to use more MCP tool calls, more agent turns, and more tokens (see also the top row of Figure~\ref{fig:pass1_vs_cost}).

Under \verb|E-Bench|-Code, however, this trend sometimes seems to reverse. With \verb|exec_code| available, models can fold multiple domain-specific function calls into a single MCP call within one turn. Still, the correlations between Avg@3 and cost are weak: Avg@3 is only weakly negatively correlated with MCP tool calls ($r=-0.36$) and is nearly uncorrelated with turns and total tokens ($r=-0.07$ and $r=-0.25$, respectively). Thus, we cannot conclude that packing more domain-specific calls into \verb|exec_code| is the only path to success, although it may increase the likelihood of solving demanding tasks efficiently.

\subsection{API Cost-Efficiency Analysis}
\label{subsec:cost_appendix}

Real deployments care not only about accuracy but also about the monetary cost of reaching it. We therefore report the task averaged API cost alongside Avg@3, under both \texttt{E-Bench} and \texttt{E-Bench}-Code in Figure~\ref{fig:price_vs_performance} and Table~\ref{tab:cost_analysis}. 
The task averaged API cost is calculated following:
Real-world deployments must balance accuracy against the monetary cost required to achieve it. We therefore report task-averaged API cost alongside Avg@3 for both \texttt{E-Bench} and \texttt{E-Bench}-Code, as shown in Figure~\ref{fig:price_vs_performance} and Table~\ref{tab:cost_analysis}. The task-averaged API cost is computed as: follows:\begin{equation}
\text{cost} = \frac{n_{\text{in}}\cdot p_{\text{in}} + n_{\text{cache}}\cdot p_{\text{cache}} + n_{\text{out}}\cdot p_{\text{out}}}{10^6},
\end{equation}
where $n_{\text{in}}$, $n_{\text{cache}}$, and $n_{\text{out}}$ are the task-average fresh-input (also, cache-activate-input), cached-input, and output tokens taken from our result files, and $p_{\text{in}}, p_{\text{cache}}, p_{\text{out}}$ are the corresponding prices in USD per million tokens (listed per model in Table~\ref{tab:cost_analysis}). 
where $n_{\text{in}}$, $n_{\text{cache}}$, and $n_{\text{out}}$ denote the number of task-averaged fresh-input (also, cache-activation input), cached-input, and output tokens, respectively; and $p_{\text{in}}, p_{\text{cache}}, p_{\text{out}}$ denote the corresponding prices in USD (\$) per million tokens for each model, as listed in Table~\ref{tab:cost_analysis}.

\begin{table}[t]
\centering
\small
\setlength{\tabcolsep}{6pt}
\renewcommand{\arraystretch}{1.1}
\caption{
    Per-task cost efficiency of the $11$ evaluated models under \texttt{E-Bench} and \texttt{E-Bench}-Code. 
    ``$p_{\text{in}}$'', ``$p_{\text{cache}}$'', and ``$p_{\text{out}}$'' are the fresh-input, cache-hit-input, and output token prices in USD per million tokens. 
    Prices for Hy3, Seed-2.1-Pro, and DeepSeek-V4-Pro are shown to four decimals because these models are priced in RMB and converted to USD manually at the current exchange rate ($1{:}6.77$); the remaining models are priced natively in USD and shown to two decimals. Rows follow the canonical \texttt{E-Bench} Avg@3 order.
}
\label{tab:cost_analysis}
\begin{tabular}{l|ccc||cc|cc}
\toprule
 & \multicolumn{3}{c||}{Price (\$/M)} & \multicolumn{2}{c|}{\texttt{E-Bench}} & \multicolumn{2}{c}{\texttt{E-Bench}-Code} \\
Model & $p_{\text{in}}$ & $p_{\text{cache}}$ & $p_{\text{out}}$ & Avg@3 & Cost (\$) & Avg@3 & Cost (\$) \\
\midrule
Kimi-K3          & 3.00   & 0.30     & 15.00  & 73.8\% & 0.634 & 77.6\% & 0.191 \\
GPT-5.5          & 5.00   & 0.50     & 30.00  & 72.0\% & 1.786 & 77.2\% & 0.481 \\
Opus-4.8         & 6.25   & 0.50     & 25.00  & 68.8\% & 3.398 & 81.1\% & 0.833 \\
Grok-4.5         & 2.00   & 0.30     & 6.00   & 66.1\% & 0.504 & 69.2\% & 0.137 \\
GLM-5.2          & 1.40   & 0.26     & 4.40   & 52.3\% & 0.258 & 61.0\% & 0.145 \\
Qwen-3.7-Max     & 1.25   & 0.25     & 3.75   & 50.9\% & 0.243 & 61.9\% & 0.138 \\
Hy3              & 0.1477 & 0.0369   & 0.5908 & 50.2\% & 0.053 & 64.4\% & 0.029 \\
Seed-2.1-Pro     & 0.8863 & 0.1773   & 4.4313 & 47.9\% & 0.452 & 53.0\% & 0.138 \\
Gemini-3.5-Flash & 1.50   & 0.15     & 9.00   & 42.6\% & 0.452 & 59.1\% & 0.189 \\
MiniMax-M3       & 0.30   & 0.06     & 1.20   & 41.1\% & 0.063 & 46.9\% & 0.041 \\
DeepSeek-V4-Pro  & 0.4350 & 0.0036   & 0.8700   & 34.5\% & 0.028 & 47.7\% & 0.024 \\
\bottomrule
\end{tabular}
\end{table}

As shown in Table~\ref{tab:cost_analysis} and Figure~\ref{fig:price_vs_performance}, API cost varies by more than an order of magnitude across models with comparable accuracy. Moreover, \texttt{exec\_code} shifts the overall frontier toward the cost-efficient upper-left region: by consolidating many domain-tool calls into compact programs, it substantially reduces per-task token usage and hence cost, while also improving Avg@3.

Under both settings, DeepSeek-V4-Pro, Hy3, Grok-4.5, and Kimi-K3 lie on the Pareto frontier of cost efficiency, achieving relatively high Avg@3 at a given API cost. From this perspective, these models offer the strongest cost-performance trade-offs among the evaluated models.

\subsection{Partial Results with Muse-Spark-1.1} \label{subsec:muse_spark_1.1}

We also tested Muse-Spark-1.1 on \verb|E-Bench| and \verb|E-Bench|-Code. However, due to its strict content-moderation mechanism, some tasks failed to be tested. For example, in \emph{Honor of Kings}, a task with narration like ``delete all that kind of friends, do not leave any of them'' consistently triggers a content-policy-violation and cannot be evaluated. Only $265$ of the $323$ tasks completed successfully across all three trials of \emph{both} settings ($268$ considering \verb|E-Bench| alone, $287$ considering \verb|E-Bench|-Code alone). We therefore compare Muse-Spark-1.1 against all other models on these $265$ succeeded tasks, re-evaluating every model on the same subset. Results are reported in Table~\ref{tab:muse_spark}.

\begin{table}[t]
\centering
\small
\setlength{\tabcolsep}{5pt}
\renewcommand{\arraystretch}{1.1}
\caption{
    Performance (Avg@3, Pass@3, Pass$^3$) and interaction cost (number of MCP tool calls and agent action turns) of the $11$ evaluated models and Muse-Spark-1.1 on the $265$ tasks that Muse-Spark-1.1 completes without content-policy-violation failure in any trial, under \texttt{E-Bench} and \texttt{E-Bench}-Code. MCP and turns are per-task averages; tokens are per-task input and output tokens in total. Rows follow the canonical \texttt{E-Bench} Avg@3 order. In each column, the best three values are highlighted in \textbf{bold}, \underline{underline}, and \colorbox{gray!20}{gray background} (highest for performance, lowest for cost).
}
\label{tab:muse_spark}
\resizebox{\textwidth}{!}{
\begin{tabular}{l||ccccc|ccccc}
\toprule
 & \multicolumn{5}{c|}{\texttt{E-Bench}} & \multicolumn{5}{c}{\texttt{E-Bench}-Code} \\
Model & Avg@3 & Pass@3 & Pass$^3$ & MCP & \#turns & Avg@3 & Pass@3 & Pass$^3$ & MCP & \#turns \\
\midrule
Kimi-K3            & \textbf{71.5\%} & \textbf{85.7\%} & \textbf{56.6\%} & 59.98 & 12.93 & \underline{75.8\%} & \underline{85.7\%} & \colorbox{gray!20}{63.8\%} & 13.32 & 9.69 \\
GPT-5.5           & \underline{70.8\%} & \colorbox{gray!20}{81.1\%} & \underline{56.6\%} & 60.39 & 12.53 & \colorbox{gray!20}{75.3\%} & \colorbox{gray!20}{85.3\%} & \underline{64.2\%} & \colorbox{gray!20}{9.31} & \textbf{7.94} \\
Opus-4.8          & \colorbox{gray!20}{69.3\%} & \underline{84.9\%} & \colorbox{gray!20}{52.5\%} & 54.48 & 11.26 & \textbf{79.9\%} & \textbf{89.8\%} & \textbf{66.0\%} & 10.01 & 9.03 \\
Grok-4.5          & 64.7\% & 79.6\% & 51.7\% & 59.04 & 30.41 & 66.7\% & 80.4\% & 53.6\% & 13.23 & 9.90 \\
\textbf{\textit{Muse-Spark-1.1}}    & 61.6\% & 75.8\% & 46.0\% & 71.59 & 15.53 & 69.9\% & 84.2\% & 53.2\% & \underline{8.79} & \underline{8.17} \\
GLM-5.2           & 52.7\% & 71.3\% & 32.1\% & 51.66 & \textbf{9.02}  & 59.4\% & 77.0\% & 39.2\% & 17.02 & \colorbox{gray!20}{8.45} \\
Hy3               & 49.6\% & 74.0\% & 26.4\% & 56.02 & 11.05 & 62.4\% & 83.4\% & 38.9\% & 20.53 & 8.58 \\
Seed-2.1-Pro      & 48.6\% & 64.5\% & 33.2\% & \colorbox{gray!20}{49.38} & 25.68 & 50.7\% & 70.2\% & 27.9\% & 11.64 & 12.11 \\
Qwen-3.7-Max      & 48.3\% & 70.9\% & 23.4\% & 50.71 & \colorbox{gray!20}{9.72}  & 60.5\% & 77.7\% & 41.5\% & 16.51 & 9.32 \\
Gemini-3.5-Flash  & 44.0\% & 64.9\% & 24.2\% & \textbf{45.41} & 10.32 & 59.1\% & 78.9\% & 36.6\% & \textbf{8.27} & 9.19 \\
MiniMax-M3        & 40.6\% & 61.9\% & 20.8\% & 52.94 & \underline{9.50}  & 45.5\% & 70.9\% & 21.1\% & 27.87 & 10.26 \\
DeepSeek-V4-Pro   & 36.5\% & 56.2\% & 19.2\% & \underline{48.80} & 10.36 & 47.5\% & 69.8\% & 26.0\% & 19.53 & 9.64 \\
\bottomrule
\end{tabular}
}
\end{table}

As shown in Table~\ref{tab:muse_spark}, Muse-Spark-1.1 is a competitive mid-frontier model that benefits clearly from code execution. On \verb|E-Bench|, it achieves $61.6\%$ Avg@3, ranking $5^{\text{th}}$ among the $12$ models and clearly outperforming the rest of the field, none of which exceeds $53\%$. With \verb|exec_code|, its Avg@3 rises to $69.9\%$, moving it to $4^{\text{th}}$. As with other models, reliability remains a challenge: Pass$^3$ is substantially lower than Pass@3. Notably, when \verb|exec_code| is available, Muse-Spark-1.1 shifts from being the most tool-intensive model---issuing $71.59$ MCP calls per task, the highest among all $12$ models, with heavy token consumption---to one of the most efficient, achieving the second-lowest number of MCP calls and agent turns per task.

Muse-Spark-1.1 is also a strong parallel tool call user. 
Under \verb|E-Bench|, Muse-Spark-1.1 issues about $4.6$ tool calls per turn ($71.59$ calls over $15.53$ turns), placing it among the strong-parallelism models alongside the frontier tier and well above the weak-parallelism outliers (Grok-4.5 and Seed-2.1-Pro, both below $2$; see Section~\ref{sec:parallel-tool-call}). Its high base call count thus reflects genuine parallel invocation within turns rather than long serial trajectories.

\section{Trajectory Case Studies}
\label{sec:trajectory-cases}

We present one Hy3 trajectory case from each of the three E-Bench domains to provide a more intuitive view of our evaluation tasks and the behaviors that the benchmark captures. Each case first shows the original Chinese user query and its English translation. Due to space limitations and for readability, the trajectory visualizations only show the names of tool calls in each round, omitting the system prompt, assistant reasoning content, and assistant content.

\newtcolorbox{trajectoryquerybox}[1]{
    enhanced,
    breakable,
    colback=blue!2!white,
    colframe=blue!35!black,
    boxrule=0.5pt,
    arc=2pt,
    left=5pt,
    right=5pt,
    top=4pt,
    bottom=4pt,
    title={#1},
    fonttitle=\bfseries\small
}

\subsection{Honor of Kings: Directional State Updates}
\label{sec:trajectory-hok}

\begin{trajectoryquerybox}{Case 1: Honor of Kings}
\begin{CJK*}{UTF8}{gbsn}
\footnotesize
\textbf{Chinese query.} 我是辩护人出击，我的ID是WX00000047。帮我整顿一下好友列表，分两件事做：

先找出最近30天里跟我一起打过对局、共同胜率不低于五成、但还没跟我建立亲密关系的好友，从里面挑亲密度最高的那个，把他设成星标好友，同时把亲密关系设成「死党」。

然后，把那些最近30天一场都没跟我打过、亲密度不到200、也没有亲密关系的好友，全部删掉。

\medskip
\textbf{English query.} I'm ``Defense Advocate Strikes'', my ID is WX00000047. Help me tidy up my friend list, in two parts:

First, find the friends who have played matches with me in the last 30 days, whose joint win rate is at least 50\%, but who have not yet established an intimate relationship with me. Among them, pick the one with the highest intimacy, set them as a starred friend, and set the intimate relationship to ``best buddy''.

Then, delete all the friends who haven't played a single match with me in the last 30 days, whose intimacy is below 200, and who have no intimate relationship with me.
\end{CJK*}
\end{trajectoryquerybox}

\begin{figure}[htbp]
    \centering
    \includegraphics[width=\textwidth]{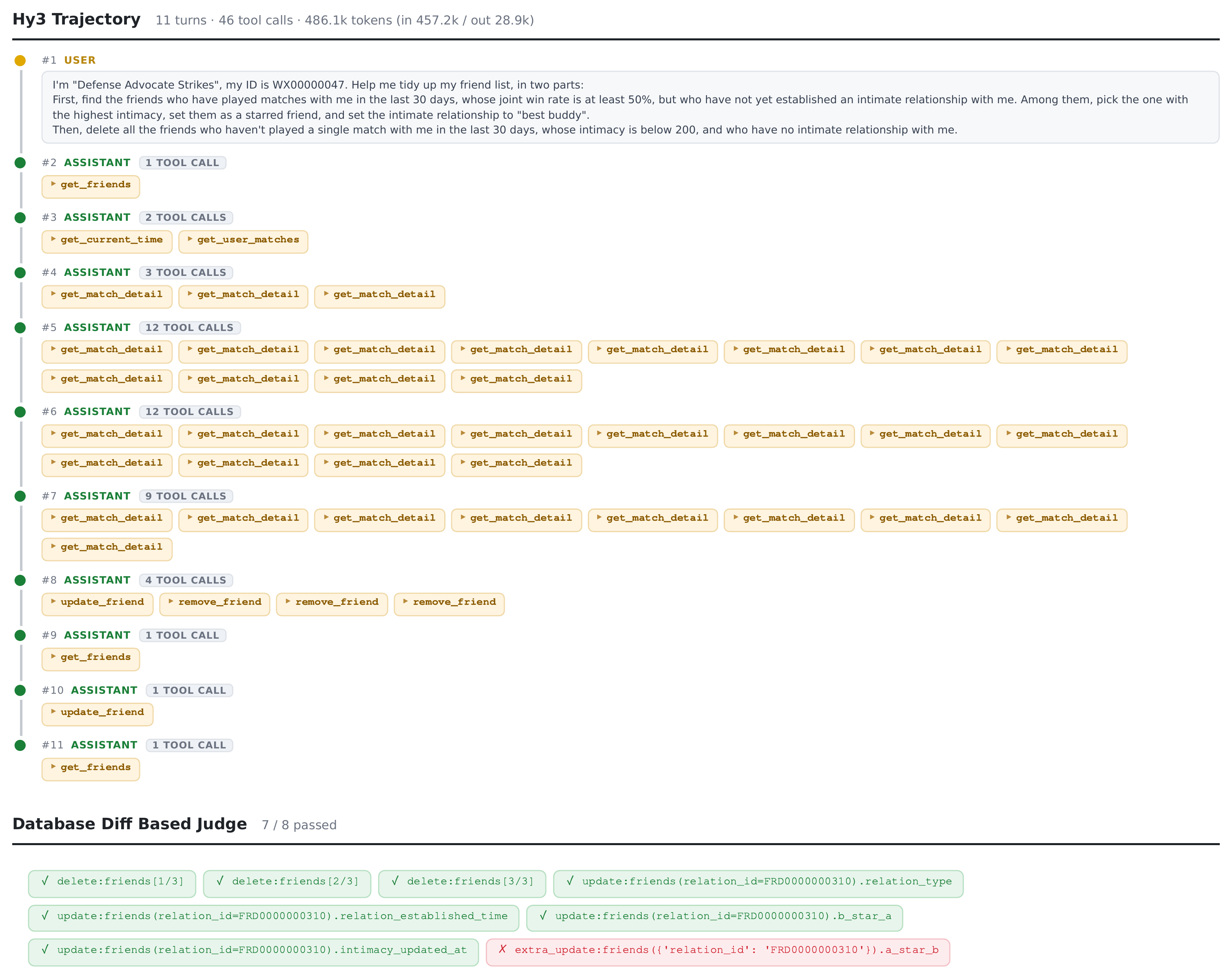}
    \caption{Hy3 trajectory on the Honor of Kings case. The task is almost completed, but one directional friend-relation field is written incorrectly and remains as an extra state change.}
    \label{fig:trajectory-hok}
\end{figure}

This failed trajectory contains 11 model calls, 10 tool-use rounds, and 46 tool calls, with a maximum of 12 parallel tool calls in one round. Hy3 satisfies 7 out of 8 checks, but fails because of one extra state update. The model observes the field \texttt{a\_star\_b} in the friend-relation tool schema and assumes that it means ``I star the other user.'' However, in this relation record the current user is on the \texttt{user\_b} side, so the correct direction is \texttt{b\_star\_a=1}. The model later re-checks the relation and writes the correct \texttt{b\_star\_a=1}, but does not clear the previously written \texttt{a\_star\_b=1}. Because E-Bench uses field-level exact state diff for verification, this unexpected extra write is sufficient to make the task fail. This case illustrates why our benchmark stresses not only finding the right entities, but also performing precise state-changing operations. Figure~\ref{fig:trajectory-hok} shows the corresponding trajectory.

\subsection{Tencent Meeting: Long-Chain Multi-Entity Orchestration}
\label{sec:trajectory-tencent-meeting}

\begin{trajectoryquerybox}{Case 2: Tencent Meeting}
\begin{CJK*}{UTF8}{gbsn}
\footnotesize
\textbf{Chinese query.} 我是市场营销部渠道运营组的胡伟杰，在深圳鹏创大厦办公。突发渠道事件，要紧急拉一个跨部门对齐小会，麻烦帮我安排一下。参会人包括我自己，再从我所在的渠道运营组里挑除我之外工龄最长的两位深圳在职正式员工，另外从产品与设计中心下面的视觉设计组挑工龄最长的那位深圳在职正式员工，一共四个人，我来当发起人和主持人。时间请在本周 4 月 21 日周一到 4 月 25 日周五之间、每天 9 点到 18 点之间，按半小时为步进扫一下这四人的日程，挑出大家共同空闲的最早一个连续 1 小时时段。会议室就在我们深圳鹏创大厦里找，要普通会议室，能装下 4 个人、容量最小的那间。腾讯会议名字就叫『【紧急】渠道-视觉对齐会』，记得加到日历并发邮件通知。会议室按上面的时间同步预定好，四位参会人也都各自建一条会议类型的日程，地点填选中的那间会议室，开启提醒并关联到这场会议。最后再建一个项目群，群名叫『渠道×视觉紧急对齐项目群』，由我从渠道运营组发起，把这四个人拉进来，群描述写『渠道运营组与视觉设计组的跨部门紧急对齐协作群，用于追踪渠道物料与视觉设计相关的协同问题，成员为双方工龄最长的核心骨干。』就行。

\medskip
\textbf{English query.} I'm Hu Weijie from the Channel Operations Group of the Marketing Department, working at Pengchuang Building in Shenzhen. There's a sudden channel incident and I need to urgently call a small cross-department alignment meeting; please help me arrange it. Attendees include myself, plus the two most senior Shenzhen-based full-time active employees from my own Channel Operations Group besides me, and also the most senior Shenzhen-based full-time active employee from the Visual Design Group under the Product \& Design Center---four people in total, with me as the organizer and host. For the time, scan these four people's schedules between Monday April 21 and Friday April 25 this week, from 9:00 to 18:00 each day, in half-hour steps, and pick the earliest continuous 1-hour slot when everyone is free. Find the meeting room within our Pengchuang Building in Shenzhen; it must be an ordinary meeting room that can hold 4 people, the one with the smallest capacity. Name the Tencent Meeting ``[Urgent] Channel-Visual Alignment Meeting'', remember to add it to the calendar and send an email notification. Book the meeting room for the same time as above; also create a meeting-type schedule entry for each of the four attendees, with the location set to the chosen meeting room, reminders enabled, and linked to this meeting. Finally, create a project group named ``Channel $\times$ Visual Urgent Alignment Project Group'', initiated by me from the Channel Operations Group, add these four people into it, and set the group description to: ``A cross-department urgent alignment collaboration group between the Channel Operations Group and the Visual Design Group, used to track collaboration issues related to channel materials and visual design; members are the most senior core backbone from both sides.''
\end{CJK*}
\end{trajectoryquerybox}

This is a representative tool-call-heavy orchestration task. The agent must identify the requester, traverse the organization hierarchy to select attendees under tenure, city, employment-status, and group constraints, scan four calendars in half-hour increments to find the earliest common one-hour slot, select the smallest eligible meeting room, and then cascade a sequence of writes: create the meeting, reserve the room, create linked calendar events for all attendees, create the project group, and add members. Hy3 completes all 15 checks with 94 tool calls across 13 model calls. The trajectory in Figure~\ref{fig:trajectory-tencent-meeting} in the main text demonstrates stable long-chain orchestration over multiple entity types, where each earlier decision constrains subsequent writes.

\subsection{QQ Music: Parallel Tool Invocation and Token Cost}
\label{sec:trajectory-qq-music}

\begin{trajectoryquerybox}{Case 3: QQ Music}
\begin{CJK*}{UTF8}{gbsn}
\footnotesize
\textbf{Chinese query.} 我是 user\_maqiang，把我听过5遍及以上、但还没收藏的歌，全都一口气加到我的收藏里。

\medskip
\textbf{English query.} I'm user\_maqiang. Add all the songs I've listened to 5 or more times but haven't favorited yet, all at once, to my favorites.
\end{CJK*}
\end{trajectoryquerybox}

\begin{figure}[htbp]
    \centering
    \includegraphics[width=\textwidth]{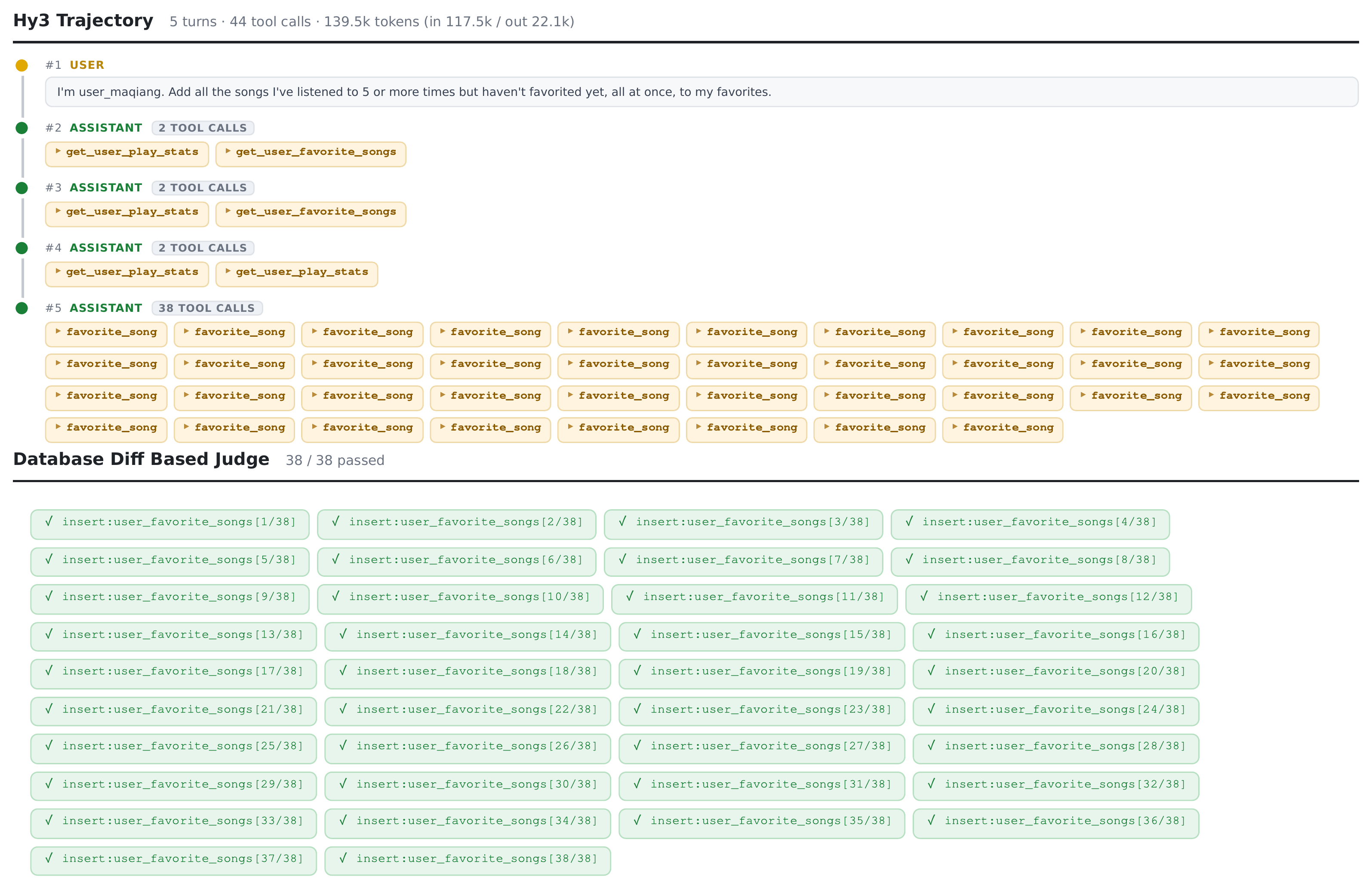}
    \caption{Hy3 trajectory on the QQ Music case. Hy3 completes the 38 favorite operations in one highly parallel round.}
    \label{fig:trajectory-qq-music-hy3}
\end{figure}

This task is semantically simple: find all songs with at least five plays that have not been favorited, and add them to the user's favorites. However, the API does not provide a batch-favorite operation, so the agent must invoke \texttt{favorite\_song} once for each selected song. Hy3 first identifies 38 target songs and then issues all 38 favorite calls in one parallel round, reaching a maximum parallelism of 38. It finishes the task in 5 model calls and 4 tool-use rounds, passing all 38 checks with 139.5K total tokens. Figure~\ref{fig:trajectory-qq-music-hy3} shows the Hy3 trajectory.

\begin{figure}[htbp]
    \centering
    \includegraphics[width=\textwidth]{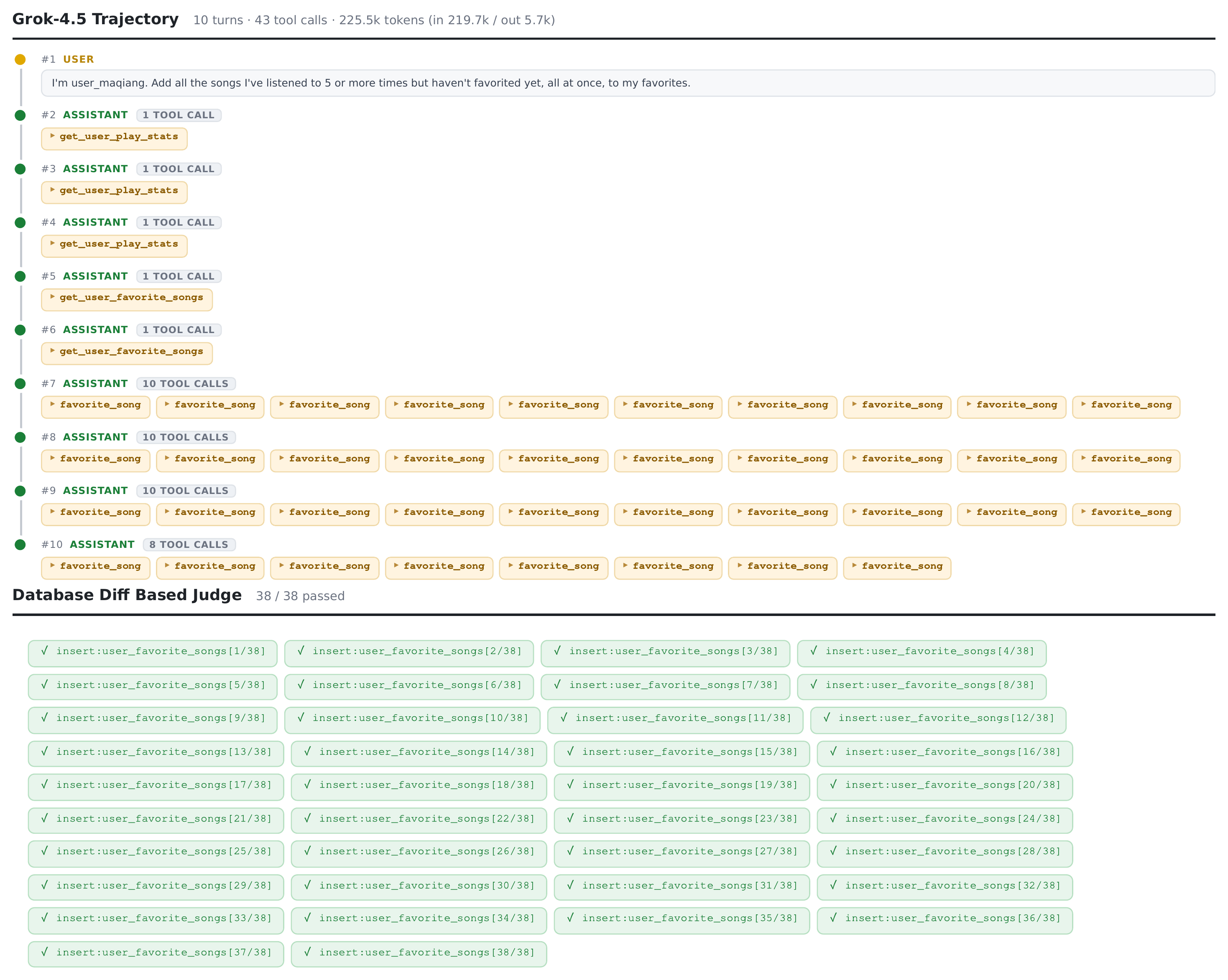}
    \caption{Grok-4.5 trajectory on the QQ Music case. Grok-4.5 executes the same favorite operations in several smaller batches and consumes substantially more tokens.}
    \label{fig:trajectory-qq-music-grok}
\end{figure}

For comparison, the corresponding Grok-4.5 trajectory in Figure~\ref{fig:trajectory-qq-music-grok} also succeeds but has much weaker parallelism: it issues at most 10 tool calls in one round and therefore splits the same favorite operations into several serial batches. This increases the trajectory to 10 model calls and 9 tool-use rounds, consuming 225.5K total tokens, about 1.6$\times$ Hy3's cost. The case provides a concrete instance of the pattern discussed in Section~\ref{sec:parallel-tool-call}: weaker parallel tool invocation leads to more rounds, repeated context re-processing, and higher token expenditure.

\end{document}